\documentclass[journal]{IEEEtran} % use the `journal` option for ITherm conference style
\IEEEoverridecommandlockouts
\usepackage{cite}
\usepackage{amsmath,amssymb,amsfonts}
\usepackage{algorithm}
\usepackage{algorithmic}
\usepackage{float} % 用于[H]位置限定
\makeatletter
\renewcommand{\ALG@name}{Algorithm} % 将"Algorithm"改为"ALGORITHM"如果需要
\makeatother
\usepackage{bm} % 加粗数学符号
\usepackage{mathrsfs}    
\usepackage{booktabs} % 在导言区添加该宏包
\usepackage{graphicx}
\usepackage{float}
\usepackage[T1]{fontenc} % 支持 \k 命令
\usepackage[utf8]{inputenc} % 支持 Unicode 输入
\usepackage{textcomp}
\usepackage{xcolor}

\usepackage{multirow}
\def\BibTeX{{\rm B\kern-.05em{\sc i\kern-.025em b}\kern-.08em
    T\kern-.1667em\lower.7ex\hbox{E}\kern-.125emX}}
\begin{document}

\title{AirLLM: Diffusion Policy-based Adaptive LoRA for Remote Fine-Tuning of LLMs over the Air}
% delete or comment-out the following line before submission
\author{
    Shiyi Yang\thanks{S. Yang, X. Yu, R. Li, and J. Zhu are with the College of Information Science and Electronic Engineering, Zhejiang University, Hangzhou 310027, China (e-mail: \{12531136; sdwhyxx; lirongpeng; zhujh20\}@zju.edu.cn).}, 
    Xiaoxue Yu, 
    Rongpeng Li, 
    Jianhang Zhu, 
    Zhifeng Zhao\thanks{Z. Zhao is with Zhejiang Lab, Hangzhou 310012, China, and also with Zhejiang University, Hangzhou 310027, China (e-mail: zhaozf@zhejianglab.com).}, 
    Honggang Zhang\thanks{H. Zhang (e-mail: honggang.zhang@ieee.org).}
}
\maketitle

\begin{abstract}
Operating Large Language Models (LLMs) on edge devices is increasingly challenged by limited communication bandwidth and strained computational and memory costs. Thus, cloud-assisted remote fine-tuning becomes indispensable. Nevertheless, existing Low-Rank Adaptation (LoRA) approaches typically employ fixed or heuristic rank configurations, and the subsequent over-the-air transmission of all LoRA parameters could be rather inefficient.
To address this limitation, we develop AirLLM, a hierarchical diffusion policy framework for communication-aware LoRA adaptation. Specifically, AirLLM models the rank configuration as a structured action vector that spans all LoRA-inserted projections. To solve the underlying high-dimensional sequential decision-making problem, a Proximal Policy Optimization (PPO) agent generates coarse-grained decisions by jointly observing wireless states and linguistic complexity, which are then refined via a Denoising Diffusion Implicit Model (DDIM) to produce high-resolution, task- and channel-adaptive rank vectors.
The two modules are optimized alternatively, with the DDIM trained under the Classifier-Free Guidance (CFG) paradigm to maintain alignment with PPO rewards.
Experiments under varying signal-to-noise ratios demonstrate that AirLLM consistently enhances fine-tuning performance while significantly reducing transmission costs, highlighting the effectiveness of reinforcement-driven, diffusion-refined rank adaptation for scalable and efficient remote fine-tuning over the air.
\end{abstract}
\begin{IEEEkeywords}
    Remote Fine-Tuning, Diffusion Policy, Reinforcement Learning, Adaptive Low-Rank Adaptation.
\end{IEEEkeywords}

\section{Introduction}
Large Language Models (LLMs) have demonstrated strong generalization across a wide range of natural language processing tasks \cite{openai2023gpt4,deepseek2024v2}.
However, their substantial parameter and memory footprints create major challenges for efficient deployment \cite{zhao2023survey}, especially on edge devices with limited computational and storage resources.
Full fine-tuning of such large models entails enormous computational and memory demands, making it infeasible for on-device learning. 
To reduce these costs, Parameter-Efficient Fine-Tuning (PEFT) \cite{lialin2023scaling} methods such as Low-Rank Adaptation (LoRA) \cite{hu2022lora} and its adaptive variant AdaLoRA \cite{zhang2023adalora} decompose updates into low-rank matrices or dynamically allocate rank budgets based on Singular Value Decomposition (SVD). While effective in reducing computational load, they primarily focus on the training process itself, often overlooking the deployment constraints of real-world systems.

\begin{figure}
    \centering
    \includegraphics[width=\linewidth]{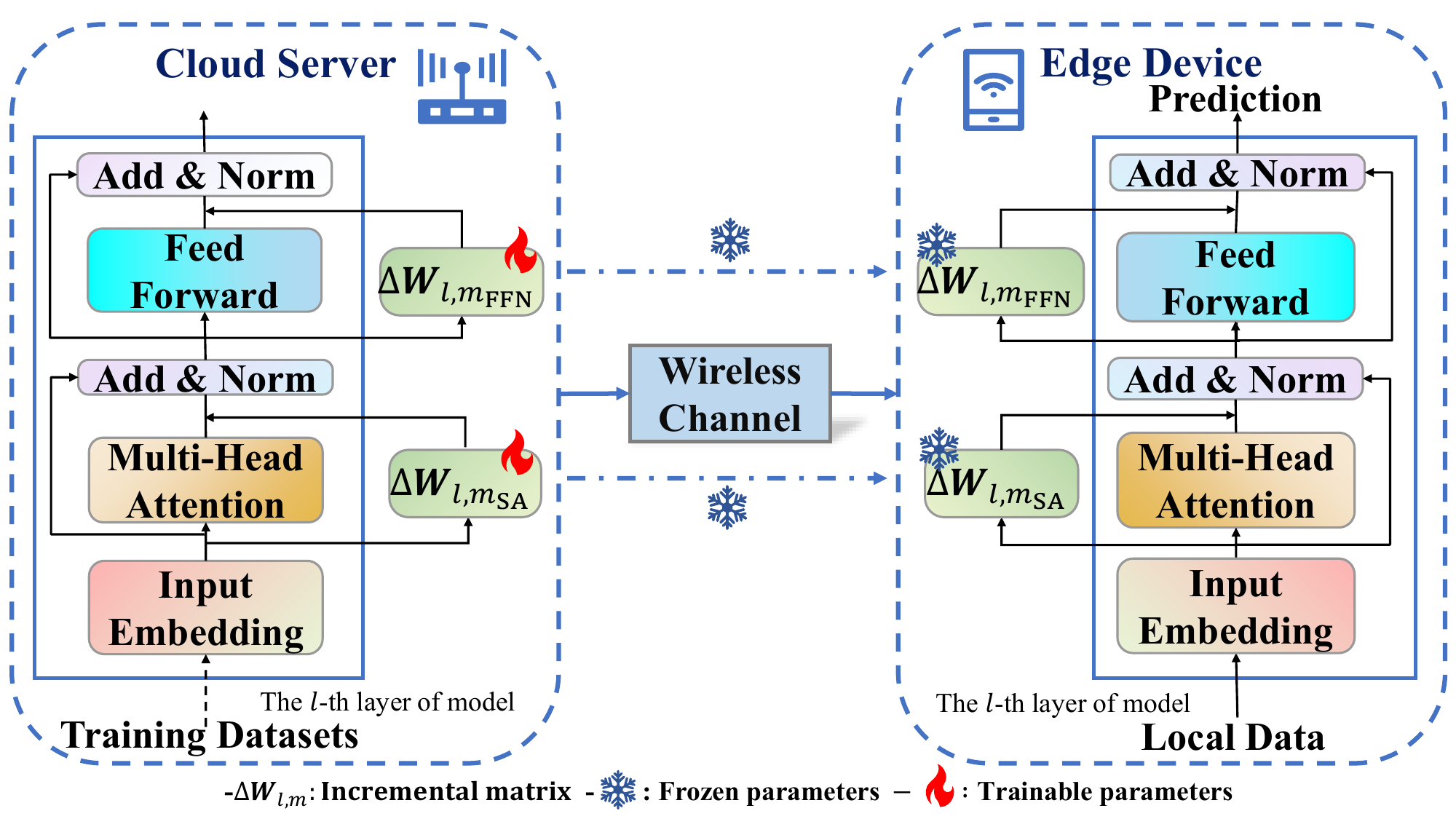}
    \caption{An illustrative example of remote fine-tuning of an LLM over the air.}
    \label{fig:application}
\end{figure}
\begin{color}{black} 
However, existing PEFT methods are poorly suited for remote fine-tuning under real-time wireless channel dynamics, since the reliance on static rank allocation and the ignorance of communication overhead leads to either redundant transmissions or insufficient performance. 
\end{color} 
\textcolor{black}{This limitation is further exacerbated in edge-cloud collaborative learning.}
\textcolor{black}{Edge devices usually lack sufficient resources for local fine-tuning, and it is difficult to meet global end-to-end constraints such as the edge device latency limit. So model adaptation is mainly performed on cloud servers \cite{hummaida2016adaptation}.}
\textcolor{black}{The updated parameters must be transmitted to the edge over wireless channels as illustrated in Fig. \ref{fig:application}, which have limited bandwidth and fluctuating signal-to-noise ratio (SNR).}
Meanwhile, training data collected from different sources often exhibits diverse characteristics and complexities, such as high Out-Of-Vocabulary (OOV) rates or varying lexical entropy \cite{xu-etal-2020-curriculum}, indicating the vocabulary mismatch and linguistic variability. These factors directly affect the relative importance of different layers during adaptation. More critically, most existing transmission and compression techniques are designed independently from tuning strategies \cite{han2016deep}, leading to suboptimal trade-offs between communication cost and adaptation quality. 
Therefore, to overcome these limitations, efficient remote fine-tuning of LLMs \cite{qiu2020pretrained} on the cloud and the transmission of partial, essential parameters to the edge emerge as a promising solution. Wireless channel configurations and propagation conditions, however, directly affect this transmission process \cite{yang2007channel}. In other words, % determining which layers to update and how much rank to allocate to each. The depth of modern LLMs significantly increases the dimensionality of the decision space, while 
only a subset of layers meaningfully contribute to downstream task performance, the allocated ranks are carefully calibrated for particular LLM projection, feed-forward network (FFN) \cite{bebis1994feed}, and output layers. 
\textcolor{black}{Unfortunately, traditional heuristics such as greedy selection or fixed allocation \cite{zhang2023adalora} cannot directly account for such dynamic, high-dimensional, and multi-objective requirements. This limitation becomes more pronounced when adaptation must track time-varying wireless channels \cite{bas2019real}.}

To address these issues, \textcolor{black}{we propose AirLLM, a communication-aware adaptive LoRA framework that dynamically adjusts layer-wise ranks based on real-time wireless channel conditions and fine-tuning dataset properties.} By formulating high-dimensional rank allocation as a sequential decision-making problem under communication constraints, AirLLM leverages Reinforcement Learning (RL) \cite{szepesvari2022algorithms} to learn adaptive rank allocation policies. The proposed RL agent observes both data and channel state information and dynamically adjusts rank budgets to balance model accuracy and transmission cost. In particular, SNR has been used as an adaptive input under dynamic channel conditions \cite{qi2025adaptive}. However, standard RL algorithms still struggle with the underlying high-dimensional action spaces. Therefore, due to high-dimensional modeling capability, we resort to the classifier-free denoising diffusion \cite{ren2024diffusionpolicy,ho2020denoising}. %Originally developed for high-dimensional control and generative modeling, diffusion policies can effectively capture dependencies across layer-wise decisions and offer improved convergence in large action spaces.
Featuring the integration of semantic-channel aware state design, diffusion-based policy generation, and a multi-objective reward formulation, our proposed framework, \textbf{AirLLM}, \textcolor{black}{resolves the mismatch between conventional PEFT methods and the practical requirements of wireless remote fine-tuning, obtaining} balanced fine-tuning performance under bandwidth constraints.

Key contributions are as follows:

\begin{itemize}
\item \textbf{Formulating Remote Fine-Tuning as a Markov Decision Process Problem}: We propose a novel formulation of remote LLM fine-tuning as a Markov Decision Process (MDP), where each decision corresponds to a layer-wise low-rank configuration under resource constraints.
\textcolor{black}{To support effective policy learning, we design a comprehensive RL state space.
It includes wireless channel statistics (e.g., SNR), data complexity metrics (e.g., lexical entropy, OOV rate) \cite{xu-etal-2020-curriculum}, and current rank assignments.}
Additionally, a multi-objective reward function is constructed to balance downstream task performance with communication efficiency, enabling the RL agent to adaptively optimize fine-tuning strategies in dynamic environments.

\item \textbf{Hierarchical Diffusion Policy for Scalable Structured Action Generation}: To address the challenge of high-dimensional decision-making in layer-wise rank allocation, we propose a hierarchical policy framework that combines a lightweight Proximal Policy Optimization (PPO)-based coarse policy with a conditional Denoising Diffusion Implicit Model (DDIM) \cite{song2020denoising}-based refinement module.
\textcolor{black}{Instead of directly predicting discrete rank configurations, the PPO agent \cite{schulman2017proximal} outputs a continuous latent vector.
This vector provides coarse layer-wise guidance for the DDIM sampler, which then refines the final structured rank vector across layers.}
Particularly, DDIM adopts the Classifier-Free Guidance (CFG) strategy to better align the denoising process with task rewards. Through training PPO and DDIM modules alternately, our approach significantly improves training stability and enables scalable action generation in large configuration spaces. % We further adapt the DDIM trajectory to accommodate structural constraints, achieving notable improvements in convergence speed and policy quality.

\item \textbf{Extensive Experimental Validation and Efficiency Gains}: We conduct comprehensive experiments across multiple wireless communication scenarios and natural language understanding tasks to evaluate the effectiveness of our framework. Results demonstrate that compared to existing PEFT baselines \cite{zhang2023adalora}, our method reduces parameter transmission costs by up to $12.5\%$ while improving task accuracy by $0.69\%$. Moreover, the proposed Diffusion-RL hybrid framework achieves over $30\%$ training efficiency gains compared to vanilla PPO \cite{schulman2017proximal}, validating its practical applicability in real-world cloud-edge deployment settings.
\end{itemize}

This paper is organized as follows. Section \ref{sec:related work} reviews related literature.
 Section \ref{sec:System Model and Problem Formulation} presents the system model and formulates the problem. Section \ref{sec:methodology}
 details the proposed hierarchical model for fine-tuning.
 Section \ref{sec:experimental_results} evaluates performance via simulations. Finally, the main conclusion is drawn in Section \ref{Conclusion}.
 
\section{Related Work}\label{sec:related work}

\subsection{Parameter-Efficient Fine-Tuning for Large Language Models}

The rapid advancement of LLMs has prompted an extensive investigation into PEFT methods, primarily motivated by the prohibitive cost of full-model updates in large-scale architectures. LoRA \cite{hu2022lora} pioneers this topic by replacing full fine-tuning with low-rank adaptation modules. Specifically, it approximates the update $\Delta \mathbf{W} \in \mathbb{R}^{m \times n}$ using two low-rank matrices $\mathbf{A} \in \mathbb{R}^{m \times r}$ and $\mathbf{B} \in \mathbb{R}^{r \times n}$, with $r \ll m, n$, significantly reducing trainable parameters while preserving downstream performance. However, LoRA and its early variants typically assign a uniform rank $r$ across all layers and modules, neglecting the inherent heterogeneity in importance and complexity among different components \cite{zhang2023adalora, clarke2024peft}.

To address this limitation, AdaLoRA \cite{zhang2023adalora} dynamically adjusts rank allocation by estimating parameter importance during training. It employs heuristic criteria derived from the SVD of parameter matrices and gradient information to guide adaptive rank adjustment. Despite these improvements, its assumption of static data distributions and system environments implies a lack of explicit mechanisms to handle dynamic and noisy wireless channels that are typical in cloud-to-edge parameter transmission \cite{zhang2023intelligent}.

\subsection{Adaptive Rank Allocation in Edge-Cloud Settings}

Despite the progress in PEFT methods for high-performance environments, little attention has been paid to communication constraints when deploying fine-tuned parameters on resource-constrained edge devices. In edge-cloud collaborative learning, fine-tuning is commonly offloaded to cloud servers, necessitating the transmission of updated low-rank parameters over wireless channels with varying bandwidth and SNR \cite{rappaport2020wireless, goldsmith2005wireless}.

Recent systems such as dLoRA dynamically orchestrate requests and LoRA adapters to improve serving efficiency \cite{wu2024dlora}. However, such serving-level optimization does not adapt fine-tuning ranks to wireless channel conditions or dataset complexity. To address this gap, we formulate adaptive rank allocation as a sequential decision-making problem and leverage RL to jointly optimize rank assignments based on dataset characteristics (e.g., lexical entropy, OOV rates) and real-time channel states (e.g., SNR, bandwidth) \cite{sutton2018reinforcement}.

\subsection{Reinforcement Learning for Rank Adaptation}

The integration of RL into PEFT for dynamic rank adaptation remains relatively under-explored. While conventional RL algorithms such as PPO \cite{schulman2017proximal} are theoretically applicable, they encounter practical challenges in large-scale Transformer-based LLMs. Specifically, assigning ranks across multiple Transformer blocks induces a high-dimensional action space, often resulting in inefficient learning and suboptimal policy performance \cite{peng2024reinforcement, heess2015learning}.

To mitigate this, latent-space policies and distributional RL have been studied for high-dimensional decision making \cite{haarnoja2018latent, lee2024off}. Diffusion policies further formulate policy generation as a denoising process for complex, structured action distributions \cite{chi2023diffusion,ren2024diffusionpolicy}. In our approach, we adopt the DDIM \cite{song2020denoising} for its accelerated inference compared to Denoising Diffusion Probabilistic Models (DDPM) \cite{ho2020denoising}, while preserving expressive power for exploration. This design enables efficient rank allocation policies that balance communication cost and model accuracy under realistic edge-cloud constraints.

\textcolor{black}{In summary, 
compared with existing works, the key differences of this paper are three-folded: First, unlike static PEFT methods \cite{hu2022lora, zhang2023adalora}, AirLLM realizes adaptive rank allocation by integrating wireless channel states and data characteristics. Second, compared with traditional RL \cite{schulman2017proximal} or regression-based refinement techniques, our hierarchical PPO-DDIM framework efficiently handles high-dimensional rank spaces and non-stationary
wireless environments. Third, extensive simulation results demonstrate that we achieve a better balance between fine-tuning accuracy and communication cost, outperforming baselines in practical remote fine-tuning scenarios.
}
\begin{table}[!t]
\caption{Main notations used in this paper.}
\label{table_notation}
\centering
\footnotesize
\setlength{\tabcolsep}{3pt}
\renewcommand{\arraystretch}{1.05}
\begin{tabular}{@{}p{0.25\columnwidth}p{0.69\columnwidth}@{}}
\toprule
\textbf{Symbol} & \textbf{Definition} \\
\midrule
$L$ & Number of Transformer layers \\
$d_h$ & Hidden dimension of the model \\
$d_s$ & Dimension of state \\
$d_{\text{MLP}}$ & Hidden size of the PPO policy multilayer perceptron (MLP) \\
$\mathbf{s}_t$ & State at step $t$ \\
$\mathbf{a}_t$ & Action vector at $t$ \\
$\mathbf{a}_t^{\text{out}}$ & Coarse-grained action output by PPO \\
$\tilde{\mathbf{a}}_t$ & Refined action after DDIM optimization \\
$R_t$ & Reward \\
$\mathcal{U}(\mathbf{r}_t)$ & Task loss under rank configuration \\
$\eta(\mathbf{r}_t)$ & Communication cost of rank configuration \\
$\lambda$ & Weighting between loss and communication cost \\
$\pi_\theta(\mathbf{a} \mid \mathbf{s})$ & PPO policy function \\
$\hat{A}_t$ & Advantage estimate \\
$J_{\text{CLIP}}(\theta)$ & PPO clipped objective \\
$\mathcal{L}_{\text{PPO}}$ & Total PPO loss \\
$\mathbf{x}_0$ & Clean rank vector (target configuration) \\
$\mathbf{x}_\tau$ & Noisy data at diffusion step $\tau$ \\
$\hat{\mathbf{x}}_0$ & DDIM-predicted clean rank vector \\
$f_\psi$ & Noise prediction network in DDIM \\
$\mathcal{L}_{\text{DDIM}}$ & DDIM loss combining noise and reward terms \\
$\kappa$ & Weighting for DDIM loss terms \\
$T_{\text{diff}}$ & Number of DDIM denoising steps \\
$r_{\text{max}}$ & Maximum rank constraint \\
$\mathbf{W}_{l,m}^{(0)}$ & Pre-trained weights at layer $l$, module $m$ \\
$\mathbf{W}_{l,m}^{(t)}$ & Fine-tuned weights at step $t$ \\
$r_{l,m}^{(t)}$ & Rank of module $(l,m)$ at step $t$ \\
$\mathscr{D}$ & Fine-tuning dataset \\
$\mathrm{SNR}_t$ & Signal-to-noise ratio \\
$W_t$ & Channel bandwidth \\
$h_t$ & Channel gain \\
$\mathbf{n}_t$ & AWGN noise \\
$C_t$ & Channel capacity \\
$H_t$ & Lexical entropy of dataset at step $t$ \\
$\rho_t$ & OOV rate of dataset at step $t$ \\
\bottomrule
\end{tabular}
\end{table}

\section{System Model and Problem Formulation}\label{sec:System Model and Problem Formulation}
\subsection{System Model}
For convenience, we also list the main notations used in this paper in Table \ref{table_notation}.

\subsubsection{LLM Parameter-Efficient Fine-tuning}
We consider a pre-trained Transformer-based LLM $\mathscr{M}$ composed of $L$ layers, where each layer $l \in \{1, \ldots, L\}$ contains a self-attention (SA) block and an FFN block. Without loss of generality, we exemplify the PEFT of an LLM by AdaLoRA \cite{zhang2023adalora}. In the SA block, AdaLoRA modules (i.e., projection matrices) are inserted into the frozen projection matrices: query ($\mathbf{W}^Q$), key ($\mathbf{W}^K$), value ($\mathbf{W}^V$), and output ($\mathbf{W}^O$). In the FFN block, AdaLoRA is similarly applied to the two linear transformations $\mathbf{W}^{\text{fc1}}$ and $\mathbf{W}^{\text{fc2}}$. During fine-tuning, the original weight matrices remain frozen, while the low-rank update matrices introduced by LoRA \cite{hu2022lora} are optimized. Mathematically, the set of LoRA-inserted modules for the $l$-th layer can be written as:
\begin{equation}
\mathcal{M}_{l} = \{ \mathbf{W}_l^Q, \mathbf{W}_l^K, \mathbf{W}_l^V, \mathbf{W}_l^O, \mathbf{W}_l^\text{fc1}, \mathbf{W}_l^\text{fc2} \}.
\label{eq:m_mods}
\end{equation}

Building upon AdaLoRA, we parameterize the weight update matrix using SVD \cite{abdi2007singular} to support dynamic and fine-grained rank allocation. For each module $m$ in layer $l$ given in Eq. \eqref{eq:m_mods}, the fine-tuned weight matrix at the $t$-th training step is expressed as:
\begin{equation}
\mathbf{W}^{(t)}_{l,m} = \mathbf{W}_{l,m}^{(0)} + \mathbf{P}^{(t)}_{l,m} \mathbf{\Lambda}^{(t)}_{l,m} \mathbf{Q}^{(t)}_{l,m},\label{eq:svd}
\end{equation}
where \textcolor{black}{$\mathbf{P}_{l,m} \in \mathbb{R}^{d_h \times r^{}_{l,m}}$} and \textcolor{black}{$\mathbf{Q}_{l,m} \in \mathbb{R}^{r_{l,m} \times d_h}$} are orthogonal matrices with the hidden dimension $d_h$, and the rank of the diagonal singular value matrix \textcolor{black}{$\mathbf{\Lambda}_{l,m} \in \mathbb{R}^{r_{l,m} \times r_{l,m}}$} with the dimension \textcolor{black}{$r_{l,m}$}. In this way, the total number of trainable parameters introduced by AdaLoRA is given by \textcolor{black}{$\sum_{l = 1}^{L} \sum_{m \in \mathcal{M}_{l}}  r_{l,m}\left(2 d_h + 1\right)$}.

To encourage the orthogonality of $\mathbf{P}_{l,m}$ and $\mathbf{Q}_{l,m}$, a \textit{regularization term} is commonly imposed in practice, namely,
\begin{color}{black}\begin{equation}
\mathcal{R} = \| {\mathbf{P}_{l,m}}^\top \mathbf{P}_{l,m} - \mathbf{I} \|_F^2 + \| \mathbf{Q}_{l,m} {\mathbf{Q}_{l,m}}^\top - \mathbf{I} \|_F^2,
\end{equation}
\end{color}
where the Frobenius norm $
\|\mathbf{A}\|_F = \sqrt{\sum_{i=1}^m \sum_{j=1}^n |a_{ij}|^2},
$ of a matrix $\mathbf{A} \in \mathbb{R}^{m \times n}$ quantifies the element-wise distance between matrices \cite{bottcher2008frobenius}. This regularization ensures that $\mathbf{P}_{l,m}$ and $\mathbf{Q}_{l,m}$ approach orthogonal matrices \cite{zhang2023adalora}. %, as $\mathbf{P}^\top\mathbf{P} = \mathbf{I}$ and $\mathbf{Q}\mathbf{Q}^\top = \mathbf{I}$ for perfectly orthogonal matrices.

On the other hand, the involved fine-tuning dataset $\mathcal{D} = \{D_1, D_2, \dots, D_N\}$ can be characterized by lexical entropy and OOV rate. Specifically, for each sample $ D_n \in \mathcal{D}$, the lexical entropy is defined as $H(D_n) = -\sum_{w \in \mathcal{V}} \hat{p}(w) \log \hat{p}(w)$, where $\hat{p}(w)$ is the empirical probability of word $w \in D_n$, and $\mathcal{V}$ is the vocabulary. Meanwhile, the OOV rate is calculated by $\rho(D_n) = \frac{|\{w \in D_n: w \notin \mathcal{V}\}|}{|D_n|}$, which measures the proportion of words in $D$ that are not in the pre-defined vocabulary $\mathcal{V}$.

\subsubsection{Wireless Communication Channel Model}
For the remote fine-tuning of LLMs, we assume a typical cloud-edge deployment setting where the fine-tuning process is conducted on the cloud server, and the resulting low-rank parameters are transmitted to resource-constrained local edge devices. %Since these devices lack the capability to perform full-scale training, the communication channel becomes a crucial bottleneck in the end-to-end system performance.
We model the wireless transmission of fine-tuned parameters from the cloud service (transmitter) to the edge device (receiver) by a slow fading Additive White Gaussian Noise (AWGN) channel \cite{goldsmith2005wireless}. Suppose each fine-tuned parameter of a module corresponding to\begin{color}{black} $\mathbf{P}_{l,m}$, $\mathbf{\Lambda}_{l,m}$ and $\mathbf{Q}_{l,m}$\end{color} is first compressed and then packed into a codeword $\mathbf{c}_t \in \mathbb{C}^{b_t}$ of length $b_t$ for transmission. The received signal at the edge is given by:
\begin{equation}
\mathbf{y}_t = h_t \mathbf{c}_t + \mathbf{n}_t,
\end{equation}
where $\mathbf{y}_t$ is the received signal at the local device, 
$h_t$ is the complex, slow-fading channel gain that remains constant over each block. 
$\mathbf{n}_t \sim \mathcal{CN}(0, \sigma_t^2 \mathbf{I}_{b_t})$ represents AWGN, with $\mathbf{I}_{b_t}$ being the $b_t$-dimensional identity matrix ensuring independence of noise elements. The notation $\mathcal{CN}$ denotes the complex normal distribution, where the mean is zero and the variance is $\sigma_t^2$. 
The average transmit power is constrained by $\frac{1}{b_t} \sum_{i=1}^{b_t} \|\mathbf{c}_t(i)\|^2 \leq P$.

Assuming perfect Channel State Information (CSI) is available at the receiver, %the effective channel is modeled as an AWGN channel with random SNR. T
the average achievable data rate over such a fading channel at the $t$-th step is:
\begin{equation}
\label{eq:channel}
C_t = \mathbb{E}_{h_t} \left[ W_t \log_2\left(1 + \text{SNR}_t \right) \right],
\end{equation}
where $W_t$ is the channel bandwidth while the value of SNR is calculated with $\text{SNR}_t = \frac{|h_t|^2 P}{\tilde{\sigma}_t^2}$.

\subsection{Problem Formulation}
Recalling the task of transmitting fine-tuned parameters from the cloud to the edge, the number of parameters to be transmitted for each module $m$ in layer $l$, as derived from Eq. \eqref{eq:svd}, is $r^{(t)}_{l,m}(2d_h + 1)$. %where $r_{l,m}$ is the rank and $d_h$ is the hidden dimension.
Accordingly, the transmission time can be calculated as
\begin{equation}
T^{(t)}_{l,m} = \frac{r^{(t)}_{l,m}(2d_h + 1) \cdot b_t}{C_t},
\end{equation}
where $b_t \equiv32$ for FP32. 
In particular, the total transmission time is usually constrained by a maximum allowable latency $T_{\text{max}}$, i.e.,
\begin{equation}
    \sum\nolimits_{l = 1}^L \sum\nolimits_{m \in \mathcal{M}_{l}} T^{(t)}_{l,m} \leq T_{\text{max}}.
\end{equation} 
The constraint implicitly restricts the rank allocation across layers. 
Moreover, due to the limited communication capacity, there also exists a bound for the total amount of transmitted parameters, i.e.,
$1 \leq r^{(t)}_{l,m} \leq r_{\text{max}}$,
where $r_\text{max}$ denotes the maximum allowable rank that satisfies the communication constraint. %It serves as an upper bound for each $r_{l,m}$, and 

Given these constraints on latency and rank capacity, the pivotal challenge of our scenario lies in effectively allocating ranks across modules and layers to maximize fine-tuning performance while adhering to communication budgets.
Notably, since computing task-specific metrics such as accuracy at every training step is computationally expensive, the loss value $\mathcal{U}(\mathbf{r}_t)$ on the fine-tuning dataset $\mathcal{D}$ is adopted as a surrogate measure to monitor intermediate performance, thereby indirectly reflecting the model's fitness under a given rank configuration $\mathbf{r}_t = \{r^{(t)}_{l,m}\}_{l = 1, m \in \mathcal{M}_{l}}^L$. 
Meanwhile, the associated communication cost $\eta(\mathbf{r}_t)$ is determined by the quantity of parameters transmitted from the cloud to the edge, which can be computed as: 
\begin{equation}
    \eta(\mathbf{r}_t) = \frac{\sum_{l = 1}^L \sum_{m \in \mathcal{M}_{l} } r_{l,m}^{(t)} (2d_h+1) \cdot b_t}{C_t\cdot T_{\max}}.
\end{equation}
Mathematically, the optimization objective can be encapsulated as
\begin{align}
\min_{\mathbf{r}}
&\quad \mathcal{U}(\mathbf{r}_{t}) + \lambda \  \eta(\mathbf{r}_t)\nonumber\\
\text{s.t.}
&\quad \sum\limits_{l = 1}^{L} \sum\limits_{m \in \mathcal{M}_{l}} \frac{r^{(t)}_{l,m}(2d_h + 1) \cdot b_t}{C_t} \leq T_{\text{max}},\label{eq:opt_problem}\\
&\quad 1 \leq r^{(t)}_{l,m} \leq r_{\text{max}}, \quad \forall l \in \{1, \dots, L\}, \forall m \in \mathcal{M}_{l},\nonumber
\end{align}
where $\lambda$ is a positive weighting factor that balances the significance of model performance and communication cost.
In this optimization problem, reducing the communication cost $\eta(\mathbf{r}_t)$ may lead to an increase in the loss function $\mathcal{U}(\mathbf{r}_t)$, and vice versa \cite{zhang2023intelligent}. 

Conventionally, AdaLoRA allocates rank budgets $\mathbf{r}_t$ across layers based on sensitivity scores and training dynamics \cite{zhang2023adalora}, whereas dLoRA optimizes request and adapter orchestration at serving time \cite{wu2024dlora}. Neither directly incorporates dataset-dependent loss $\mathcal{U}(\mathbf{r}_{t})$ and wireless communication capability $C_t$ into fine-tuning rank allocation. Moreover, local layer-wise decisions may overlook global interactions among layers and fail to satisfy strict end-to-end constraints on parameter transmission or inference latency. 
 Hence, identifying a meticulous design of the rank parameters for each layer that satisfies all constraints while minimizing the objective function is of utmost importance. 

\section{Methodology}\label{sec:methodology}
\subsection{RL for Dynamic Rank Allocation}
To enable dynamic rank allocation in real-world deployment scenarios, RL, particularly the PPO \cite{schulman2017proximal}, offers a promising solution for learning adaptive policies through continuous interaction with the environment. 
By adopting PPO, we seek to derive a preliminary policy that effectively balances fine-tuning accuracy and communication cost under varying system conditions. 

\subsubsection{MDP Formulation for Rank Allocation}
\begin{figure*}[!htb]
    \centering
    \includegraphics[width=0.9\linewidth]{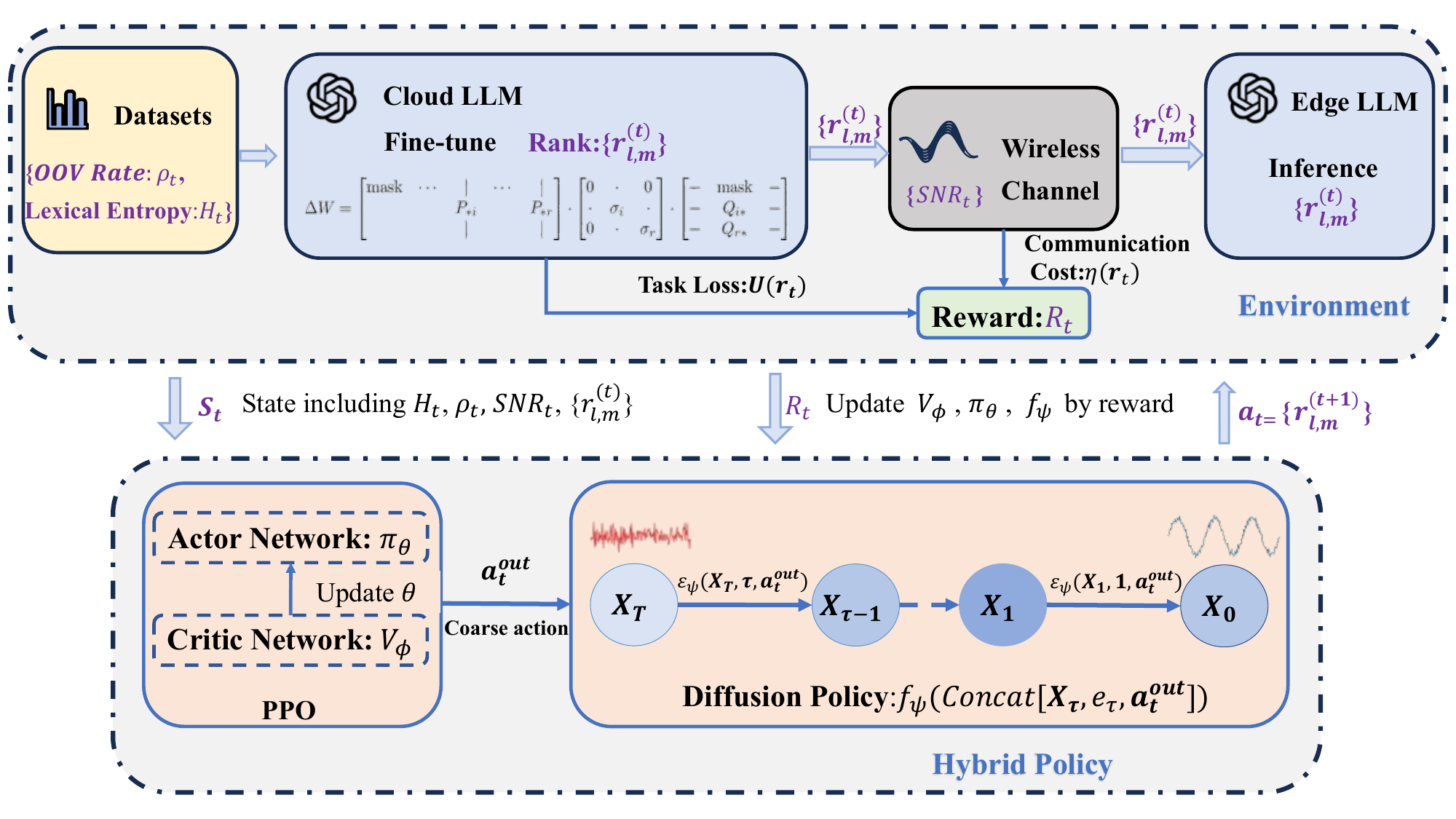}
    \caption{Diffusion policy framework for dynamic rank allocation in remote fine-tuning.}
    \label{fig:framework}
\end{figure*}

RL provides a dynamically adaptive decision-making paradigm for remote fine-tuning of large models, with its core advantage lying in real-time strategy optimization through environmental feedback. To enable adaptive rank configuration under real-world constraints such as wireless bandwidth and task complexity, we formulate the rank allocation as an MDP $(\mathcal{S},\mathcal{A},R)$. Such assumption allows the agent to learn context-aware policies that dynamically balance compression and performance, where the action directly corresponds to the rank configuration $\mathbf{r}_t$.

\begin{itemize}
\item \emph{State Representation:}
The state $\mathbf{s}_t =\{\mathrm{SNR}_t, W_t, H_t, \rho_t, \mathbf{r}_t\} \in \mathcal{S}$ is designed to comprehensively capture both the system-level constraints and task-specific complexity. Correspondingly, wireless transmission capability is implicitly acquired via $\mathrm{SNR}_t$ and $W_t$ through Eq. \eqref{eq:channel}. %This reflects the maximum number of parameters that can be feasibly transmitted in edge deployment. 
Lexical entropy $H_t$ and OOV rate $\rho_t$ characterize the complexity of the fine-tuning dataset at time step $t$, reflecting how much expressive capacity (i.e., rank) the model may require and how many tokens that do not appear in the pretraining vocabulary.
% :
% $
% H_t = - \sum_{w \in \mathcal{V}_t} p_t(w) \log p_t(w),
% $
% where $\mathcal{V}_t$ is the vocabulary set at step $t$, and $p_t(w)$ is the empirical probability of word $w$ in the dataset at $t$. 
% OOV rate
% $
% \rho_t = \frac{|\mathcal{O}_t|}{|\mathcal{D}_t|},
% $
% where $\mathcal{O}_t$ denotes the set of tokens at step $t$ that do not appear in the pretraining vocabulary, and $|\mathcal{D}_t|$ is the total number of tokens in the fine-tuning data at $t$.
 The current rank configuration $\mathbf{r}_t$, which represents the rank parameters for all modules in all layers, serves as the foundation upon which the next action (rank configuration) will be applied.

\item \emph{Action Space:}
In our formulation, each Transformer layer contains $6$ LoRA-inserted projection matrices. Thus, the action vector $\mathbf{a}_t$ at time step $t$ (which directly defines the rank configuration) is given by
$
\mathbf{a}_t \in \{0, 1, \cdots, r_{\text{max}}\}^{6L}. 
$ And the new rank distribution $\mathbf{r}_{t+1}=\mathbf{a}_t$.
% where $r_{t,l,m}$ denotes the discrete rank value for the $m$-th module in the $l$-th Transformer layer.
% The $6L$-dimensional structural action space, corresponding to the complete rank configuration, introduces two fundamental challenges for RL. First, conventional policy optimization methods such as PPO often assume a factorized distribution over action dimensions, which struggles to capture the inherent dependencies among low-rank components. Second, computing policy gradients in such high-dimensional discrete spaces can be computationally intensive, particularly when modeling joint distributions or enforcing structured constraints, with the complexity potentially scaling as $\mathcal{O}((6L)^3)$ \cite{wang2023challenges, heess2015learning}.

\item \emph{Reward Function:}
Consistent with the objective function in Eq. \eqref{eq:opt_problem}, the reward function can be written as
\begin{equation}
R_t = -\mathcal{U}(\mathbf{r}_t) - \lambda \cdot \eta(\mathbf{r}_t). %\frac{\sum_{l = 1}^L \sum_{m \in M_{mods}} r_{t,l,m} (2d_h+1)}{\mathcal{C}},
\label{eq:reward}
\end{equation}
\begin{color}{black}
where \begin{color}{black}
the cross-entropy loss $\mathcal{U}(\mathbf{r}_t)$ penalizes inaccurate predictions, pushing the policy to retain expressive model capacity, while normalized communication cost term $\eta(\mathbf{r}_t)$, which is proportional to the total number of transmitted parameters, encourages the policy to minimize unnecessary rank allocation. 
\end{color}
In addition, the coefficient $\lambda$ controls the trade-off between accuracy and efficiency. This design allows the RL agent to learn policies that adaptively allocate ranks depending on task difficulty and system bandwidth, thereby achieving communication-efficient fine-tuning without sacrificing too much performance.
\end{color}
\end{itemize}
RL aims to maximize cumulative rewards through interaction with the environment. Specifically, an agent tries to learn a policy $\pi(\mathbf{a}|\mathbf{s})$, a probability distribution over actions $\mathbf{a}$ given state $\mathbf{s}$, which guides the agent's future behaviors.

\subsubsection{PPO for Coarse-Grained Policy Optimization}
\label{sec:ppo_background}
%In sequential decision-making, RL formulates the problem as an MDP, where an agent interacts with the environment to maximize cumulative rewards. Central to this is the policy $\pi(\mathbf{a}|\mathbf{s})$, a probability distribution over actions $\mathbf{a}$ given state $\mathbf{s}$, which determines the agent's behavior. Evaluating and improving the policy relies on value-based concepts foundational to algorithms like PPO.
As noted above, we adopt PPO to train a feasible policy for dynamic rank allocation.
Beforehand, we briefly present the key ingredients related to PPO. 
In particular, the \textit{return} at time $t$ is defined as $\tilde{R}_t = \sum_{k=0}^\infty \gamma^k R_{t+k}$, where $\gamma \in (0,1)$ discounts future rewards. The \textit{state-value function} $V_\phi(\mathbf{s}_t) = \mathbb{E}_\pi[\tilde{R}_t | \mathbf{s}_t]$ estimates the expected return from $\mathbf{s}_t \in \mathbb{R}^{d_s}$, parameterized by a neural network with weights $\phi$.
The \textit{action-value function} $Q_\pi(\mathbf{s}_t, \mathbf{a}_t) = \mathbb{E}_\pi[\tilde{R}_t | \mathbf{s}_t, \mathbf{a}_t]$ quantifies the return for action $\mathbf{a}_t$ in $\mathbf{s}_t$. The \textit{advantage function} $\hat{A}_t = Q_\pi(\mathbf{s}_t, \mathbf{a}_t) - V_\phi(\mathbf{s}_t)$ measures how much better or worse $\mathbf{a}_t$ is compared to the average action at $\mathbf{s}_t$, thus guiding policy improvement. Since $Q_\pi$ is typically unavailable, PPO estimates $\hat{A}_t$ via Generalized Advantage Estimation (GAE) \cite{schulman2017proximal}, using Temporal-Difference (TD) errors:
\begin{equation}    
    \hat{A}_t = \sum\nolimits_{k=0}^\infty (\gamma \zeta)^k \delta_{t+k},\label{eq:advantage_function}
\end{equation}
where $\delta_t = R_t + \gamma V_{\phi_{\text{old}}}(\mathbf{s}_{t+1}) - V_{\phi_{\text{old}}}(\mathbf{s}_t)$ and $V_{\phi_{\text{old}}}(\cdot)$ denotes the value function under previous parameters for stable bootstrapping. $\zeta \in [0,1]$ is the decay factor of GAE, which controls the bias-variance trade-off in advantage estimation.
Eq. \eqref{eq:advantage_function} enables efficient estimation from sampled trajectories without computing full returns.

To stabilize policy updates of the $\theta$-parametrized actor, PPO further introduces a clipped objective on top of the value function loss, which can be formulated as
\begin{align}
    & {J}_{\text{CLIP}}(\theta) =
\min \left(
r_t(\theta) \hat{A}_t,
\operatorname{clip}\big(r_t(\theta), 1-\varrho, 1+\varrho\big) \hat{A}_t
\right),\nonumber\\
& J_{V}(\phi) = \left(V_{\phi}(\mathbf{s}_{t}) - \big(\hat{A}_t+V_{\phi_{\text{old}}}(\mathbf{s}_{t})\big)\right)^2,\label{eq:ppo_loss}
\\
& \mathcal{L}_{\text{PPO}} = \mathbb{E}\left[{J}_{\text{CLIP}}(\theta) + J_{V}(\phi)\right],\nonumber
\end{align}
where $r_t(\theta) = \pi_\theta(\mathbf{a}_t|\mathbf{s}_t)/\pi_{\theta_{\text{old}}}(\mathbf{a}_t|\mathbf{s}_t)$ is the probability ratio, measuring the likelihood of taking certain actions under the policy $\pi_\theta(\mathbf{a}_t|\mathbf{s}_t)$ versus the old one $\pi_{\theta_{\text{old}}}(\mathbf{a}_t|\mathbf{s}_t)$.
$\operatorname{clip}(r_t(\theta), 1-\varrho, 1+\varrho)$ denotes a clipping function that limits the deviation $r_t(\theta)$ between successive policy updates within the range $(1-\varrho, 1+\varrho)$.
By minimizing the loss in Eq. \eqref{eq:ppo_loss} via gradient descent, we optimize both the policy $\pi_\theta$ and value function $V_\pi$ in PPO to achieve improved decision performance.
The clipped surrogate objective of PPO limits abrupt policy updates, which is desirable for coarse-grained decision-making under varying channel and task states. In the proposed hierarchical design, PPO provides coarse guidance, while the subsequent refinement stage handles the high-dimensional rank configuration.

At each decision step, the PPO agent, implemented as a two-layer multi-layer perceptron (MLP), maps the current state $\mathbf{s}_t \in \mathbb{R}^{d_s}$ to a latent action prior $\mathbf{a}_t \in \{0, 1, \ldots, r_{\max}\}^{6L}$. Given a hidden width of $d_{\text{MLP}}$, the computational complexity of single forward or backward pass is $\mathcal{O}\left(d_{\text{MLP}} \cdot \big(6L \cdot (r_{\max}+1)+d_s\big) + 6L \cdot (r_{\max}+1)\right)$, where the first term accounts for the two-layer MLP inference that generates logits, and the second term corresponds to the element-wise softmax computation across the discrete action heads.
The complexity scales linearly with the input and output dimensions.

However, PPO's performance tends to degrade in high-dimensional discrete action spaces \cite{henderson2018deep, schulman2017proximal}. 
Moreover, the gradient signal becomes increasingly sparse while the reward variations across consecutive actions remain negligible, making the estimated advantage $\hat{A}_t$ highly sensitive to noise and thereby resulting in unstable policy updates \cite{andrychowicz2021matters}.

\subsection{Diffusion Policy for High-Dimensional Actions}
\label{sec:diffusion-policy}
% Instead of directly modeling the complex global data distribution, diffusion models decompose the problem into a sequence of simpler local conditional distributions, enabling the model to learn incremental structural features at varying noise levels.
% The detailed procedure is illustrated in Figure~\ref{fig:framework}.

To mitigate the challenges of PPO in high-dimensional action space, we propose to restrict PPO to generating coarse priors $\mathbf{a}_t^{\text{out}}$ only, while delegating fine-grained decision-making $\mathbf{\tilde{a}}_t$ to a conditional diffusion model. \textcolor{black}{Compared to traditional refinement techniques
like Gaussian Process Regression (GPR) \cite{williams2006gaussian}, DDIM is more appropriate for our task due to three-folded reasons: it can better capture multimodal optimal rank distributions without
GPR's Gaussian assumption, scales efficiently in high
dimensional spaces, and is robust to non-stationary wireless channels via iterative denoising rather than noise-sensitive one-step regression in GPR.}
The hierarchical coarse-to-fine strategy leverages the probabilistic noising--denoising process of diffusion models to provide a principled alternative for high-dimensional action modeling by decomposing the decision-making problem into a series of tractable local conditional distributions \cite{ren2024diffusionpolicy,ho2020denoising}. 
Through the progressive denoising procedure, the hierarchical model is able to capture structural patterns across noise scales, thereby enhancing exploration capabilities and improving robustness when combined with PPO \cite{ren2024diffusionpolicy}. The detailed procedure is illustrated in Fig. \ref{fig:framework}.

\subsubsection{Conditional Diffusion Refinement Guided by PPO Priors}

% This makes them particularly advantageous for overcoming key limitations of PPO in high-dimensional action spaces, including vanishing gradient signals and unstable policy updates resulting from high-dimensional discrete action spaces.
% Specifically, diffusion models circumvent the difficulty of directly modeling complex global data distributions , allowing the model to progressively learn structural patterns across different noise scales.

% PPO generates discrete rank configurations $\mathbf{a}_t \in \{0, 1, \ldots, r_{\max}\}^{6L}$ across all $L$ Transformer layers, leading to a joint policy over $6L$ dimensions. This results in a worst-case computational complexity of $\mathcal{O}((6L)^3)$, making it infeasible for large-scale models such as OPT-1.3B with $L=24$. Through the hierarchical coarse-to-fine strategy, PPO only generates a coarse-grained rank vector $\mathbf{a}_t^{\text{out}}$ conditioned on the current state $\mathbf{s}_t$, which is then refined by a conditional DDIM to yield the final configuration $\mathbf{\tilde{a}}_t$. This decoupled design significantly reduces the PPO-related complexity to $\mathcal{O}(d_{\text{MLP}}^2)$, where $d_{\text{MLP}}$ denotes the hidden dimension of the PPO policy network. A detailed derivation of the complexity analysis is provided in Section~\ref{sec:complexity}.

\begin{algorithm}[!t]
\caption{The Training of Hierarchical PPO-DDIM Adaptive Rank Allocation Policy.}
\label{alg:airllm}
\vspace{1mm}
\textbf{Input:} Initialized model and hyper-parameters. \\
\textbf{Output:} Trained hierarchical policy $(\pi_\theta, f_\psi)$ for adaptive rank allocation. \\
% \textbf{Initialize:} PPO policy $\pi_\theta$, DDIM model $\epsilon_\phi$, value function $V_\psi$, replay buffer $\mathcal{B}$.
\begin{algorithmic}[1]
\FOR{episode = $1, 2, \dots$,}
    \STATE Observe initial state $\mathbf{s}_0$;
    \FOR{step $t = \{1, \dots, T_\text{env}\}$}
        \STATE Generate coarse action $\mathbf{a}_t^{out} \sim \pi_{\theta_{\text{old}}}(\cdot|\mathbf{s}_t)$; 
        \STATE \begin{color}{black} // PPO outputs low-dimensional guidance \end{color}
        \STATE Initialize noise latent $\mathbf{x}_{T_{\text{diff}}} \sim \mathcal{N}(\mathbf{0}, \mathbf{I})$; 
        \FOR{$\tau = \{T_{\text{diff}}-1, \dots, 0\}$}
            \STATE Obtain the conditional noise prediction $\tilde{\mathbf{\boldsymbol{\epsilon}}}_{\tau, \psi}$ with Eq. \eqref{eq:conditional_eps}
            \STATE \begin{color}{black} // DDIM refines via $a_{t}^{out}$ conditioning\end{color}
            \STATE Predict the clean sample $\hat{\mathbf{x}}_0$ with Eq. \eqref{eq:mu_theta};
            \STATE Refine action using DDIM procedure with Eq. \eqref{eq:ddim_sample};
        \ENDFOR
        \STATE Decode refined action through Eq. \eqref{eq:final_action};
        \STATE Deploy the rank configuration to obtain the next state $\mathbf{s}_{t+1}$;
        \STATE Compute reward using Eq. \eqref{eq:reward} and collect transition $\{\mathbf{s}_t, \tilde{\mathbf{a}}_t, R_t, \mathbf{s}_{t+1}\}$ into batch $\mathcal{B}$;
    \ENDFOR
    \STATE Compute PPO loss using samples from $\mathcal{B}$ with Eq. \eqref{eq:ppo_loss};
    \STATE Update PPO policy $\pi_\theta$ and value function $V_\pi$;
    % \STATE Sample mini-batch $\{(\mathbf{s}_j, \mathbf{a}_j, R_j)\}$ from $\mathcal{B}$;
    \FOR {$\tau = \{0, \dots, T\}$}
        \STATE Generate noisy sample $\mathbf{x}_\tau$ from clean input $\mathbf{x}_0$ using  Eq. \eqref{eq:forward_diffusion};%补充
        \STATE Compute the hybrid loss according to Eq. \eqref{eq:hy_loss};
        \STATE Update the parameters in noise network $f_\psi$ via the gradient descent.
    \ENDFOR
\ENDFOR
\end{algorithmic}
\end{algorithm}

We begin by briefly introducing the forward and reverse processes of the DDPM and its deterministic variant DDIM, which achieves faster inference with similar generative quality \cite{song2020denoising}. Notably, DDIM shares the same training objective and noise schedule as DDPM, with the main distinction residing in the sampling procedure used during inference.

\paragraph{Forward Diffusion Process.}
To enable diffusion-based training, it is essential to apply a forward noise injection process to the training data. Specifically, each clean rank configuration vector $\mathbf{x}_0$ is progressively corrupted by injecting Gaussian noise over $T$ discrete steps. Formally, at each timestep $\tau \in \{1, \dots, T\}$, the conditional distribution of the noisy latent variable $\mathbf{x}_\tau$ given the previous state $\mathbf{x}_{\tau-1}$ is defined as
\begin{equation}
q(\mathbf{x}_\tau | \mathbf{x}_{\tau-1}) = \mathcal{N}\left(\mathbf{x}_\tau; \sqrt{\alpha_\tau} \mathbf{x}_{\tau-1}, (1 - \alpha_\tau)\mathbf{I} \right),
\label{eq:forward_diffusion}
\end{equation}
where the noise schedule parameters $\alpha_\tau \in (0,1)$ control the amount of noise injected, and $\mathcal{N}(\cdot; \bm{\mu}, \bm{\Sigma})$ denotes a Gaussian distribution with mean $\bm{\mu}$ and covariance $\bm{\Sigma}$. By marginalizing over intermediate variables, the closed-form distribution of $\mathbf{x}_\tau$ conditioned on the clean sample $\mathbf{x}_0$ is given by
\begin{equation}
q(\mathbf{x}_\tau | \mathbf{x}_0) = \mathcal{N}\left(\mathbf{x}_\tau; \sqrt{\bar{\alpha}_\tau} \mathbf{x}_0, (1 - \bar{\alpha}_\tau)\mathbf{I} \right).
\end{equation}
% \begin{equation}
%      \bar{\alpha}_\tau = \prod_{s=1}^\tau \alpha_s= \prod_{s=1}^t (1 - \beta_s).\label{eq:beta}
% \end{equation}
Here, $\bar{\alpha}_\tau = \prod_{s=1}^\tau \alpha_s= \prod_{s=1}^t (1 - \beta_s)$ represents the cumulative product of the noise schedule up to timestep $\tau$.

\paragraph{Conditional Reverse Denoising Process.}
The reverse diffusion process aims to iteratively denoise $\mathbf{x}_\tau$ back to the clean signal estimate $\mathbf{x}_0$. Serving as a critical component of the inference procedure, the denoising step is parameterized by a neural network $f_\psi(\cdot)$ that predicts the injected noise at each step. To incorporate guidance from the PPO output $\mathbf{a}_t^{\text{out}}$, the noise estimator is conditioned on both the noisy latent and the PPO output $\mathbf{a}_t^{\text{out}}$, together with the timestep embedding $\mathbf{e}_\tau$ that encodes the current diffusion step.
Mathematically, the obtained conditional noise prediction, denoting as $\tilde{\boldsymbol{\epsilon}}_{\tau, \psi} $, can be encapsulated as
\begin{equation}
\tilde{\boldsymbol{\epsilon}}_{\tau, \psi} = f_\psi \left( \text{Concat}[\mathbf{x}_\tau, \mathbf{e}_\tau, \mathbf{a}_t^{\text{out}}] \right), % (\mathbf{x}_\tau, \tau, \mathbf{a}_t^{\text{out}} = f_\psi 
\label{eq:conditional_eps}
\end{equation}
where $\mathbf{e}_\tau$ is typically implemented as a sinusoidal positional encoding to provide temporal context, and $f_\psi(\cdot)$ denotes the denoising network, which can be realized as a U-Net architecture \cite{rombach2022high} or a residual MLP depending on model scale and complexity.

Using the predicted noise, the posterior mean estimator of the clean signal at timestep $\tau$ is computed as
\begin{equation}
\mu_\psi(\mathbf{x}_\tau, \tau, \mathbf{a}_t^{\text{out}}) = \frac{1}{\sqrt{\alpha_\tau}} \left( \mathbf{x}_\tau - \frac{1 - \alpha_\tau}{\sqrt{1 - \bar{\alpha}_\tau}}  \tilde{\boldsymbol{\epsilon}}_{\tau, \psi} \right).
\label{eq:mu_theta}
\end{equation}
Following the standard DDPM posterior mean, the expression can incorporate the PPO prior as a conditioning signal and generate a fine-tuned $\tilde{\mathbf{a}}_t$ for the rank allocation task.

\paragraph{Deterministic Sampling via DDIM.}
For inference, we employ the DDIM sampler, which enables deterministic and efficient reconstruction of the clean action vector. Starting from an initial latent $\mathbf{x}_T \sim \mathcal{N}(0, \mathbf{I})$, the reverse trajectory is computed recursively, where at each step $\tau-1$, the latent vector $\mathbf{x}_{\tau-1}$ is estimated from the current value $\mathbf{x}_\tau$ as
\begin{equation}
\mathbf{x}_{\tau-1} = \sqrt{\bar{\alpha}_{\tau-1}}  \hat{\mathbf{x}}_0 + \sqrt{1 - \bar{\alpha}_{\tau-1} - \sigma_\tau^2} \tilde{\boldsymbol{\epsilon}}_{\tau, \psi} + \sigma_\tau \boldsymbol{\epsilon},
\label{eq:ddim_sample}
\end{equation}
where $\hat{\mathbf{x}}_0 := \mu_\psi(\mathbf{x}_\tau, \tau, \mathbf{a}_t^{\text{out}})$ is the predicted clean sample, $\boldsymbol{\epsilon} \sim \mathcal{N}(0, \mathbf{I})$ is an additional noise term, and $\sigma_\tau \geq 0$ controls the stochasticity of the sampler with $\sigma_\tau=0$ yielding fully deterministic DDIM sampling.% This scheme balances deterministic reconstruction fidelity and exploration diversity.

After completing $T_{\text{diff}}\leq T$ reverse steps, the final latent $\mathbf{x}_0$ is transformed into the actionable discrete vector by rounding and clipping, which is formulated by
\begin{equation}
\tilde{\mathbf{a}}_t = \mathrm{clip}\left( \lfloor \mathbf{x}_0 \rfloor, \, 0, \, r_{\max}  \right),
\label{eq:final_action}
\end{equation}
where $r_{\max}$ denotes the maximum allowed rank value per dimension, ensuring validity and executability of the generated action. $\lfloor \cdot \rfloor$ represents element-wise flooring to the nearest lower integer.

\subsubsection{Hybrid Training Objective and Optimization.}
To enhance controllability without relying on external classifiers, the training adopts the CFG strategy \cite{ho2022classifierfreediffusionguidance}.
Specifically, the conditional diffusion model is trained to minimize a hybrid loss that combines reconstruction fidelity in noise prediction and policy effectiveness measured via task reward, 
\begin{equation}
\mathcal{L}_{\text{DDIM}} =
\mathbb{E}_{\mathbf{x}_0, \tau, \boldsymbol{\epsilon}, \mathbf{a}_t^{\text{out}}} 
\left[ \left\| \tilde{\boldsymbol{\epsilon}}_{\tau, \psi} - \boldsymbol{\epsilon} \right\|_2^2 \right] - \kappa \cdot \mathbb{E}_{\mathbf{s}_t, \tilde{\mathbf{a}}_t}[R_t],\label{eq:hy_loss}
\end{equation}
where the noise $\boldsymbol{\epsilon} \sim \mathcal{N}(0, \mathbf{I})$ is the ground-truth noise added at timestep $\tau$, $R_t$ is the reward obtained from executing the refined action $\tilde{\mathbf{a}}_t$ in state $\mathbf{s}_t$  as defined in Eq. \eqref{eq:reward}, and $\kappa > 0$ is a hyperparameter balancing denoising accuracy and policy reward maximization.

To maintain stability and promote complementary learning between the coarse and fine policies, we adopt an alternating training strategy that iteratively updates PPO and DDIM with shared trajectory data.
Algorithm \ref{alg:airllm} summarizes the entire inference and training process for reference.

%\begin{itemize}
%    \item \textbf{PPO Stage:} The PPO policy $\pi_\theta$ is trained using environment trajectories $\{(\mathbf{s}_t, \mathbf{a}_t, R_t, \mathbf{s}_{t+1})\}$ to optimize the clipped surrogate loss in Eq.~\eqref{eq:ppo_loss}. The immediate reward $R_t$ is used to estimate returns $\tilde{R}_t$ and advantage signals $\hat{A}_t$, which guide updates to both the actor $\pi_\theta$ and critic $V_\phi$. The resulting output 
    % $\mathbf{a}_t^{\text{out}} = \pi_\theta(\mathbf{s}_t)$
%    $\mathbf{a}_t^{\text{out}} \sim \pi_\theta(\cdot \vert \mathbf{s}_t)$
%    serves as a coarse action prior.

%    \item \textbf{DDIM Stage:} Holding $\pi_\theta$ fixed, we train the DDIM-based fine policy $f_\psi$ to refine $\mathbf{a}_t^{\text{out}}$ into fine-grained executable actions $\tilde{\mathbf{a}}_t$. Specifically, we sample clean actions $\mathbf{x}_0 = \tilde{\mathbf{a}}_t$, perturb them to obtain noisy latents $\mathbf{x}_\tau$, and optimize the hybrid denoising loss , as defined in Eq.~\eqref{eq:hy_loss}.
%\end{itemize}

\begin{color}{black}Notably, the inference follows a clear two-stage design that tightly couples PPO and DDIM. Fist, PPO observes the wireless channel state (including SNR and bandwidth) and the linguistic complexity of the dataset, then outputs a low-dimensional coarse-grained guidance vector $\textbf{a}_t^{out}$. This vector captures the overall trade-off between fine-tuning accuracy and communication cost, providing stable high-level guidance for rank allocation. Based on this conditional guidance, DDIM further explores the structural dependencies among Transformer layers, and expands the coarse guidance into a complete high-dimensional, layer-specific rank configuration $\tilde{\textbf{a}}_t$. In this way, PPO focuses on global decision-making, while DDIM performs a fine-grained optimization to maintain model performance.

To ensure stable and effective training, we adopt an alternating training strategy. In the PPO update stage, we fix the DDIM module and use the obtained reward $R_t$ to update the PPO policy via the clipped surrogate loss (Eq.~\eqref{eq:ppo_loss}), so that PPO gradually learns better coarse-grained guidance strategies.
In the DDIM update stage, we fix the PPO module and use high-quality trajectories from the replay buffer to train DDIM.
DDIM learns to map the coarse guidance to high-performance rank configurations by minimizing the hybrid loss function (Eq.~\eqref{eq:hy_loss}). Such alternating optimization ensures that both components reinforce each other and steadily improve the overall performance. 
\end{color}
\subsubsection{Computational Complexity Analyses}
\label{sec:complexity}
\begin{color}{black}
To circumvent the severe inefficiency of vanilla PPO in high-dimensional discrete action spaces, we reformulate the policy as a low-dimensional continuous vector generator and discard the costly softmax operation, cutting down redundant complexity by $\mathcal{O}(6L \cdot (r_{\max}+1))$.
Different from vanilla PPO suffering from exponential complexity in the $6L$-dimensional discrete action space, our revised lightweight PPO outputs only a compact low-dimensional latent vector $\mathbf{a}_t^{\text{out}}$.
Assuming the dimension of this coarse latent representation is $D_{\text{low}} \ll 6L$, the computational complexity of our modified PPO policy network scales as $\mathcal{O}(D_{\text{low}}^2)$, which is negligible and nearly constant with respect to the full $6L$ space.
Notably, we decompose the decision-making process into two complementary stages: coarse-grained PPO guidance and fine-grained DDIM refinement.
The coarse-grained output $\mathbf{a}_t^{\text{out}}$ produced by PPO has a tiny dimension, far smaller than that of the final $6L$-dimensional discrete rank vector ${\mathbf{a}}_t \in \mathbb{Z}^{6L}$.
Such dimensional decoupling is the core of our complexity optimization and effectively eliminates the curse of dimensionality.

The fine-grained action optimization is completed by a conditional DDIM module, which takes the coarse latent $\mathbf{a}_t^{\text{out}}$ as conditional guidance, and refines a randomly initialized Gaussian latent $\mathbf{x}_T \sim \mathcal{N}(0, \mathbf{I})$ into a reasonable structured rank vector.
For DDIM equipped with a two-residual-block MLP denoiser with hidden width $d_{\text{latent}}$, its per-step denoising complexity scales linearly with $6L$ and $d_{\text{latent}}$.
Given that $T_{\text{diff}}$ (diffusion steps) is set to a small fixed constant, the total inference complexity of DDIM is $\mathcal{O}(T_{\text{diff}} \cdot L \cdot d_{\text{latent}})$, keeping efficient and tractable even for large-scale Transformer models with huge $L$.
When the more powerful U-Net is adopted as the denoiser, the per-step complexity is $\mathcal{O}\left( C_{\text{in}}^2 \cdot L_s \cdot 2^{L_{\text{depth}}} \right)$ where $C_{in}$ denotes the input channel dimension of U-Net, $L_s=6L$ represents the dimension of the rank action space, and 
$L_{depth}$ is the network depth of U-Net. After $T_\text{diff}$ denoising steps, the total inference cost turns to $\mathcal{O}\left( T_\text{diff} \cdot C_{\text{in}}^2\cdot L_s \cdot 2^{L_{\text{depth}}} \right)$. This cost still follows linear scaling with $L_s$ and avoids exponential growth, remaining feasible for deployment.

In a nutshell, the total inference complexity of the hybrid PPO-DDIM framework sums the overhead of the low-dimensional PPO branch and the diffusion refinement branch. The modified PPO introduces only negligible $\mathcal{O}(D_{\text{low}}^2)$ computational overhead, while the DDIM component scales linearly with the target high-dimensional action size $6L$. This hybrid design thoroughly avoids the exponential complexity of vanilla PPO in high-dimensional spaces, guaranteeing efficient and stable inference for large-scale LLM low-rank adaptation.
\end{color}

\section{Experimental Results \label{sec:experimental_results}}
\subsection{Experimental Settings}
This section presents comprehensive experimental results designed to validate the effectiveness of AirLLM for remote fine-tuning of LLMs. %Our experiments systematically evaluate key components, including reward balancing, diffusion prediction strategies, learning rate dynamics, rank scheduling, and channel robustness. The aim of these experiments is to test whether each contributes to the predicted balance between task accuracy and transmission efficiency. 
All experiments use the SST-$2$ dataset \cite{socher2013recursive} (i.e., a binary classification dataset \textcolor{black}{for sentiment analysis, consisting of $67,349$ training samples and $872$ test samples, where each sample is a sentence labeled as positive or negative sentiment}). \textcolor{black}{We adopt the OPT-1.3B model \cite{zhang2022opt}, a decoder-only Transformer model with $24$ layers, $12$ attention heads per layer, a hidden dimension of $2,048$, and a total of $1.3$ billion parameters; this model is equipped with LoRA adapters for all linear projection layers (including attention query, key, value, and output layers) with a default rank of $r=8$ to enable parameter-efficient fine-tuning. For LLM fine-tuning, we use a batch size of $32$, weight decay of $10^{-5}$, and AdamW optimizer with $\beta_1=0.9$ and $\beta_2=0.999$.  Hyperparameters for PPO training and the diffusion policy (e.g., learning rates, diffusion steps) are fully provided in Table \ref{settings}.} Wireless communication is modeled as an AWGN channel with $100$~MHz bandwidth, $1$s latency, and SNR levels ranging from $-5$~dB to $15$~dB. The total number of training steps is set to $15,000$ and we evaluate accuracy every $100$ step during the training process. If the reward value exceeds $\min[-\lambda,\,-0.5]$ and the accuracy does not improve for $5$ consecutive evaluations, we terminate training, and such an \textit{early stopping} strategy applies for all models to ensure consistency \cite{Raskutti2014earlystopping}. Notably, this choice of reward convergence threshold is motivated by the following observation: when $\lambda$ is very small (e.g., $\lambda=0.1$ or $0.01$), the second term in Eq.~\eqref{eq:reward} has a slight influence on the overall reward, and a reward value of $-0.5$ is considered sufficient to indicate convergence. In contrast, when $\lambda$ is larger (e.g., $\lambda=1$), the second term becomes more dominant, and the overall reward is lower. In this case, we consider $-\lambda$ as a sufficiently high reward value. We compare AirLLM against AdaLoRA \cite{zhang2023adalora} and random rank allocation baselines, \textcolor{black}{where AdaLoRA is a state-of-the-art parameter-efficient fine-tuning method that adaptively allocates low-rank budgets across layers and modules (with the same LoRA rank $r=8$ as AirLLM for fair comparison), and Random Rank Allocation is a naive baseline that randomly assigns ranks to each LoRA adapter without adaptive optimization.} And we evaluate the performance in terms of binary classification accuracy.
Finally, we summarize the default parameter settings in Table \ref{settings}.
%Key hyperparameters are systematically varied, including reward balancing coefficient (\(\ze'ta\) in the range $0.01--1.0$), diffusion prediction modes (\(\epsilon\)-prediction, \(v\)-prediction, and \(x_0\)-prediction), learning rates (e.g., PPO $0.0001--0.001$ and diffusion $0.0001--0.001$), training-inference step combinations (e.g., ($1000, 50$)), and different \(\beta\) schedules (scaled-linear, cosine, linear). Channel robustness is further tested to assess AirLLM's adaptability to varying SNR conditions.
\begin{table}[!t]
\centering
\footnotesize
\caption{Default parameter settings in simulations.}
\label{settings}
\setlength{\tabcolsep}{4pt}
\renewcommand{\arraystretch}{1.05}
\begin{tabular}{@{}p{0.48\columnwidth}p{0.46\columnwidth}@{}}
\toprule
\textbf{Parameter} & \textbf{Value} \\
\midrule
MLP hidden size ($d_{\text{MLP}}$)  & $256$                  \\
Discount factor ($\gamma$)          & $0.95$                 \\
Decay factor of GAE ($\zeta$)          & $0.95$                 \\
U-Net depth ($D$)                    & $6 $                   \\
Kernel size ($k$)                   & $3 $                   \\
Batch size                          & $32$                   \\
Learning rate of PPO                      & $1 \times 10^{-4}$   \\
Learning rate of DDIM                   &          $5 \times 10^{-5}$       \\
Random seeds                        & $42$                    \\
Hybrid loss weight ($\kappa$)       & $0.1 $                 \\
Reward balance factor ($\lambda$)
  & $0.01$, $0.1$, $1$\\
Total training steps
& $15,000$\\
DDIM inference steps ($T_\text{diff}$)
& $30$, $50$, $100$\\
DDPM Training steps 
& $600$, $800$, $1,000$\\
PPO clipping factor ($\varrho$)
& $0.2$\\
Evaluating steps
& $100$\\
Maximum rank constraint $(r_{\text{max}})$ & $8$, $16$, $32$, $64$, $128$\\
SNR (dB) & $-5$, $0$, $5$, $10$, $15$ \\
DDIM prediction strategy & $\boldsymbol{v}$-, $\boldsymbol{\epsilon}$-, $\boldsymbol{x_0}$-prediction\\
Noise schedule ($\beta$) & linear, cosine, scaled-linear\\
\bottomrule
\end{tabular}
\end{table}
\begin{table*}[t]
\centering
\footnotesize
\caption{Performance comparison in terms of prediction accuracy.}
\label{tab:method_accuracy_horizontal}
\setlength{\tabcolsep}{5pt}
\begin{tabular}{c *{6}{c}}
\toprule
\multirow{ 2}{*}{\textbf{Metric} }& \multicolumn{5}{c}{\textbf{AirLLM}}   & \multirow{2}{*}{\textbf{AdaLoRA}} \\
\cline{2-6}
& \textbf{PPO} & \textbf{DDIM} & \textbf{PPO+DDPM} & \textbf{PPO+DDIM/MLP} & \textbf{PPO+DDIM/U-Net} &  \\
\midrule
Accuracy    & $94.50 \%$       & $94.72 \%$    & $95.41 \%$          & $95.53 \% $       & $95.30\%$        & $95.18 \% $          \\
\textcolor{black}{Communication Cost} & \textcolor{black}{$0.862$} & \textcolor{black}{$0.858$} & \textcolor{black}{$0.821$} & \textcolor{black}{$\mathbf{0.795}$} & \textcolor{black}{$0.808$} & \textcolor{black}{$0.843$} \\
\textcolor{black}{Convergence Steps} & \textcolor{black}{$12500$} & \textcolor{black}{$8800$} & \textcolor{black}{$9200$} & \textcolor{black}{$\mathbf{7800}$} & \textcolor{black}{$8500$} & \textcolor{black}{$8000$} \\
\textcolor{black}{Inference Latency (ms)} & \textcolor{black}{$0.95$} & \textcolor{black}{$1.02$} & \textcolor{black}{$1.28$} & \textcolor{black}{$\mathbf{1.10}$} & \textcolor{black}{$5.35$} & \textcolor{black}{$0.88$} \\
\bottomrule
\end{tabular}
\end{table*}
\subsection{Simulation Results}
\subsubsection{Performance Comparison}

\begin{figure}[t]
    \centering
    \includegraphics[width=\linewidth]{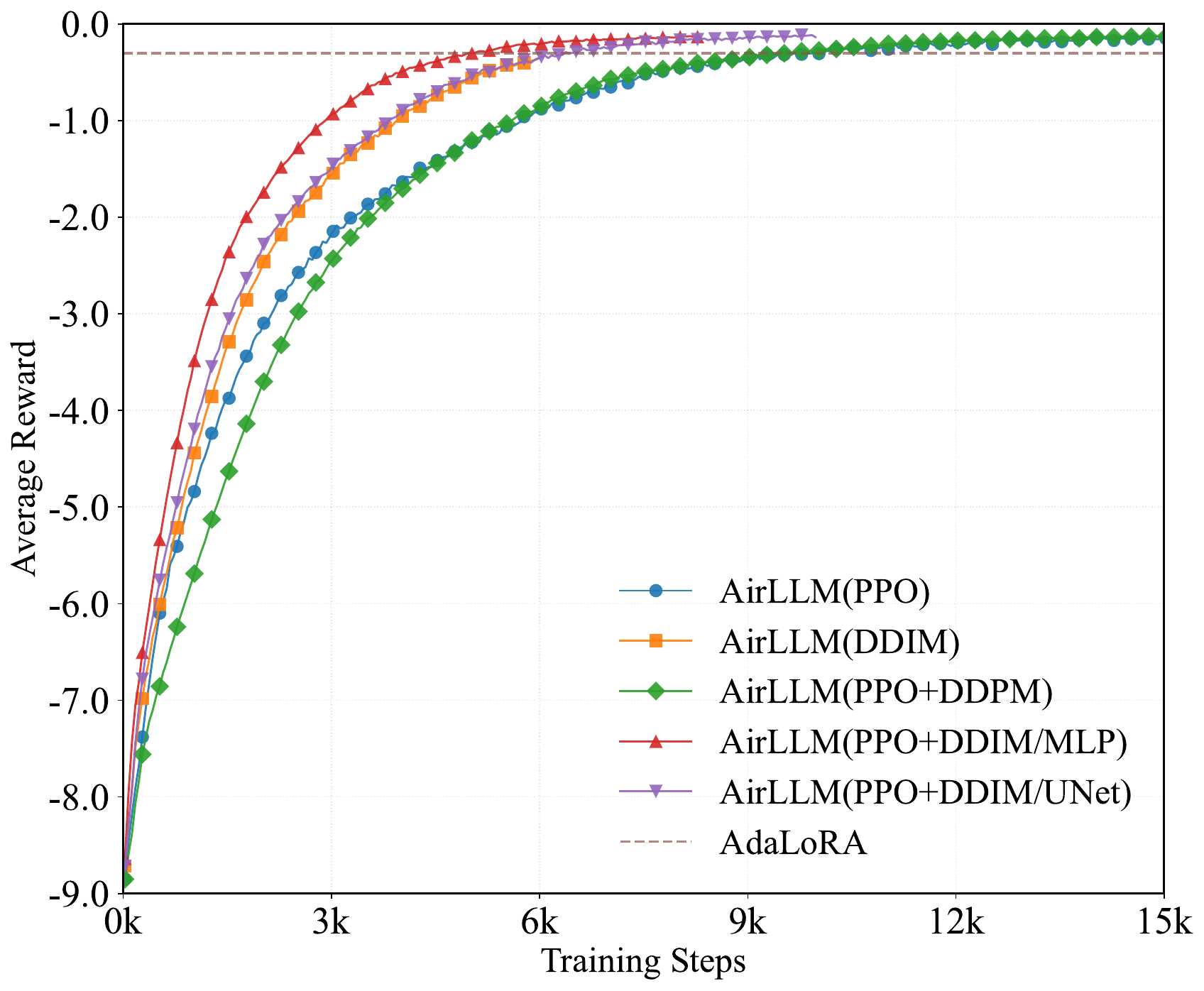}
    \caption{Comparison of average reward for different training methods.}
    \label{fig:convergence}
\end{figure}

\begin{figure}[t]
    \centering
    \includegraphics[width=\linewidth]{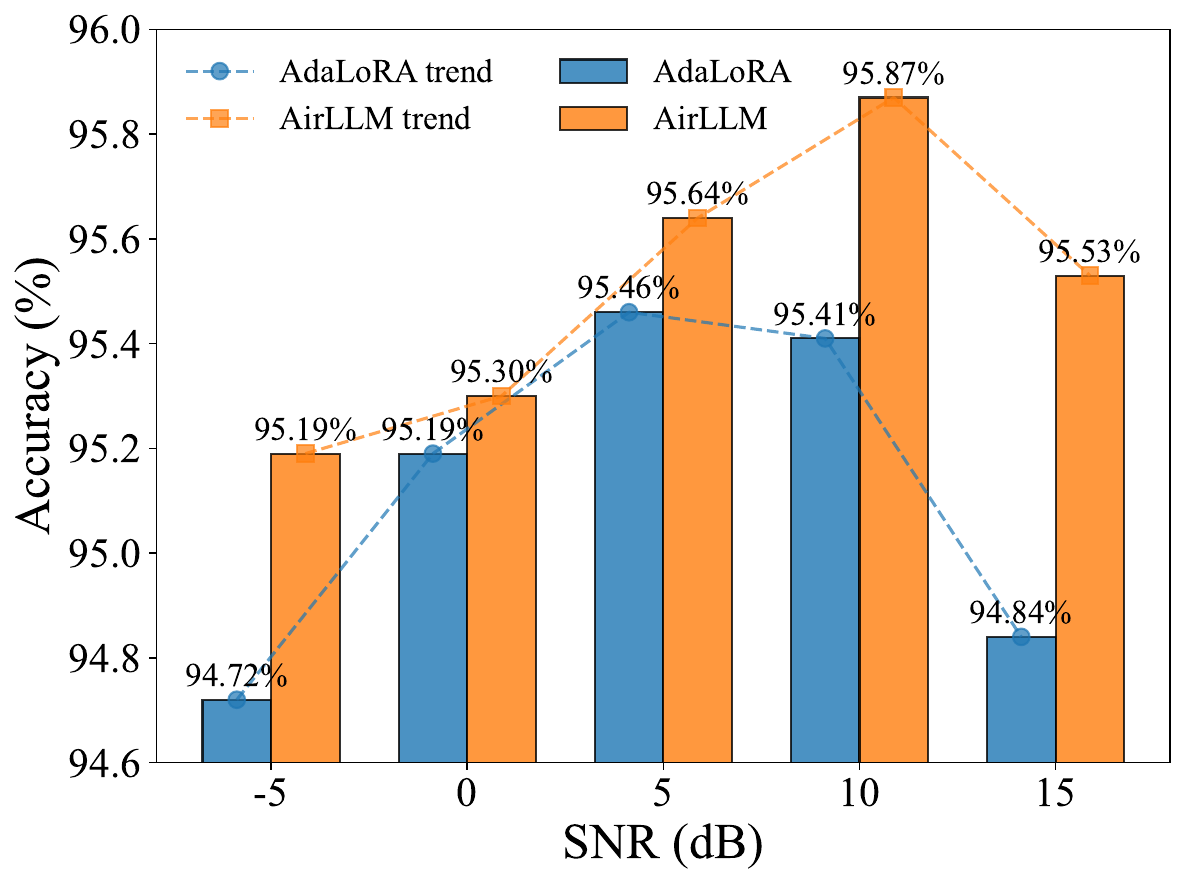}%结果图之间的字体大小差别有点大
    \caption{Accuracy performance evaluated under diverse SNRs.}
    \label{fig:snr}
\end{figure}
Table \ref{tab:method_accuracy_horizontal} comprehensively compares the proposed AirLLM (with different variants) with AdaLoRA, covering not only prediction accuracy but also \textcolor{black}{communication cost, convergence speed, and inference latency}---key metrics to evaluate the balance between performance and efficiency. Several variants of AirLLM with different RL algorithms are leveraged, and we also study the performance differences when using MLP and U-Net as backbones of DDIM. On the other hand, Fig. \ref{fig:convergence} gives the corresponding convergence curves. It can be observed from Table \ref{tab:method_accuracy_horizontal} and Fig. \ref{fig:convergence} that standalone PPO or DDIM yield the lowest accuracy, \textcolor{black}{higher communication cost, slower convergence, and less efficient inference}. \textcolor{black}{Specifically, standalone PPO and DDIM exhibit notably higher communication overhead and slower convergence compared to integrated AirLLM variants, while their inference efficiency, though relatively competitive in some cases, comes at the cost of significantly compromised accuracy. This confirms the inefficiency of single-module designs in balancing performance and deployment efficiency.} Instead, regardless of the backbone of either MLP or U-Net, the hierarchical integration achieves superior performance in all metrics. Meanwhile, the integration of PPO with DDIM accelerates the convergence compared to that with DDPM, \textcolor{black}{reduces communication overhead, and maintains moderate inference latency}. \textcolor{black}{Compared to PPO+DDPM, the PPO+DDIM combination shows clear advantages in convergence speed and communication efficiency, while keeping inference latency at a manageable level that meets the requirements of remote fine-tuning scenarios.} Finally, alongside its computational efficiency, the use of MLP also leads to better results, as shown by its lower inference latency and higher accuracy compared to U-Net-based variants in Table \ref{tab:method_accuracy_horizontal}. \textcolor{black}{Notably, MLP-based AirLLM variants maintain efficient inference latency that is comparable to baseline methods, while U-Net-based variants suffer from significantly higher latency due to their more complex network structure.} 
%Table \ref{tab:method_accuracy_horizontal} first compares the proposed AirLLM with AdaLoRA, while several variants of AirLLM with different RL algorithms are also leveraged. Notably, we also study the performance differences when using MLP and U-Net as backbones of DDIM. On the other hand, Fig. \ref{fig:convergence} gives the corresponding convergence curves. It can be observed from Table \ref{tab:method_accuracy_horizontal} and Fig. \ref{fig:convergence} that standalone PPO or DDIM yield the lowest accuracy. Instead, regardless of the backbone of either MLP or U-Net, the hierarchical integration achieves superior performance.
%Meanwhile, the integration of PPO with DDIM accelerates the convergence than that with DDPM. Finally, alongside its computational efficiency, the use of MLP also leads to better results. 
\par
\textcolor{black}{
Although AirLLM(PPO) and AirLLM(PPO+DDIM/MLP) achieve similar final rewards (i.e., a composition of task loss and communication cost) in Fig. \ref{fig:convergence}, since the task loss is not strictly proportional to classification accuracy, the more expressive diffusion model can better explore the rank space and discover more favorable LoRA rank assignments, leading to higher accuracy in Table \ref{tab:method_accuracy_horizontal}.}
% The convergence curves in Fig. \ref{fig:convergence} and accuracy comparison in (b) illustrate key insights:  
% \begin{itemize}
%  \item {Combination Necessity}:  
% Standalone PPO and DDIM show lower accuracy. Their integration (e.g., PPO+DDIM+MLP/U-Net) achieves better performance, proving combining these frameworks is critical for accuracy.  
%  \item {Convergence Speed}:  
% PPO-DDIM hybrid methods converge faster than PPO or DDIM alone. Leveraging DDIM's deterministic sampling reduces redundant exploration in high-dimensional action spaces, accelerating stable reward attainment vs. standalone PPO/DDIM or DDPM-based baselines.  
%  \item {MLP Versus U-Net}:  
% Among hybrids, PPO+DDIM+MLP outperforms PPO+DDIM+U-Net in accuracy.  
% \textit{U-Net} is a convolutional network with encoder-decoder architecture,which is often used for dense prediction tasks like image reconstruction. Here, its more complex skip connections and hierarchical feature learning introduce extra computational overhead.  
% In contrast, MLP's simpler feedforward structure enables more efficient parameterization for this task, yielding better results.  
% \end{itemize}
% In short, combining PPO and DDIM boosts both accuracy and convergence speed, with MLP as a more effective auxiliary component here.
\par
\begin{color}{black}
    As shown in Table \ref{tab:method_accuracy_horizontal}, AirLLM (PPO+DDIM/MLP) achieves $95.53\%$ accuracy, which is higher than the state-of-the-art baseline AdaLoRA ($95.18\%$). Meanwhile, our method obtains a lower communication cost of $0.795$ compared with $0.843$ for AdaLoRA, and converges much faster with only $7,800$ steps. The results confirm that AirLLM can achieve better accuracy with lower communication overhead and higher efficiency than AdaLoRA, making it more suitable for wireless remote fine-tuning.
\end{color}

In addition to efficiency and convergence speed, as shown in Fig. \ref{fig:snr}, AirLLM demonstrates superior robustness under varying channel conditions. 
%Fig. \ref{fig:snr} presents accuracy across SNR levels ranging from $5$ dB to $15$ dB, simulating practical wireless environments. 
Specifically, AirLLM achieves $0.11$-$0.69$ percentage points higher accuracy compared to AdaLoRA, consistently adapting its rank configurations under varying wireless channel conditions. 
Notably, the accuracy exhibits moderate fluctuations across different SNR conditions rather than a strictly monotonic trend. This is because AirLLM jointly optimizes task loss and communication cost rather than classification accuracy alone, and different SNR conditions may therefore lead to different rank configurations.
%This robustness is attributed to the hierarchical framework's context-aware state encoding (integrating SNR, lexical entropy, and OOV rate) and its ability to exploit DDIM's structured exploration for fine-grained adaptation.
Furthermore, benchmarking experiments in Table \ref{tab:performance_comparison} comprehensively evaluate AirLLM against AdaLoRA under varying rank budgets (i.e., $r_{\text{max}}$). The results demonstrate that AirLLM consistently outperforms the baseline across all settings. Particularly, it achieves a notable higher accuracy at $r_{\text{max}}=64$ while simultaneously reducing transmitted parameters by $12.5\%$. %This efficiency gain arises from AirLLM's dynamic, communication-aware rank allocation, where PPO provides coarse-grained policy guidance and DDIM refines rank configurations to match real-time channel conditions and model sensitivity.
These results collectively validate the effectiveness of AirLLM: by unifying PPO's stability and DDIM's high-dimensional modeling, the framework meets the dual objectives of high accuracy and communication efficiency in remote fine-tuning scenarios, confirming the theoretical motivation proposed in Section \ref{sec:methodology}.

\textcolor{black}{These results collectively validate that AirLLM achieves a superior balance between classification accuracy, communication overhead, convergence efficiency, and inference complexity, which is critical for communication-constrained remote LLM fine-tuning.}

\begin{table}[!t]
\centering
\caption{Comparison in terms of accuracy and transmitted parameters. }
\label{tab:performance_comparison}
\renewcommand{\arraystretch}{1.2}
\begin{tabular}{@{}cccrc@{}}
\toprule
\multirow{2}{*}{\begin{tabular}[m]{@{}c@{}}Rank\\ ($r_{\text{max}}$)\end{tabular}} & 
\multicolumn{2}{c}{Accuracy (\%)} & 
\multicolumn{2}{c}{Transmitted Parameters} \\
\cmidrule(lr){2-3} \cmidrule(lr){4-5}
 & AdaLoRA & AirLLM & AdaLoRA & AirLLM \\
\midrule
$8$ & $95.18$ & $\mathbf{95.64}\uparrow$ & $4,718,592$ & $4,235,648\downarrow$ \\
$16$ & $94.72$ & $\mathbf{95.30}\uparrow$ & $9,437,184$ & $8,471,296\downarrow$ \\
$32$ & $95.41$ & $\mathbf{95.87}\uparrow$ & $18,874,368$ & $16,942,592\downarrow$ \\
$64$ & $94.84$ & $\mathbf{95.53}\uparrow$ & $32,399,048$ & $28,234,160\downarrow$ \\
$128$ & $94.72$ & $\mathbf{95.19}\uparrow$ & $63,870,464$ & $56,664,832\downarrow$ \\
\bottomrule
\end{tabular}
\end{table}

\subsubsection{Performance Sensitivity Studies}

\begin{figure}[tb]
    \centering    \includegraphics[width=\linewidth]{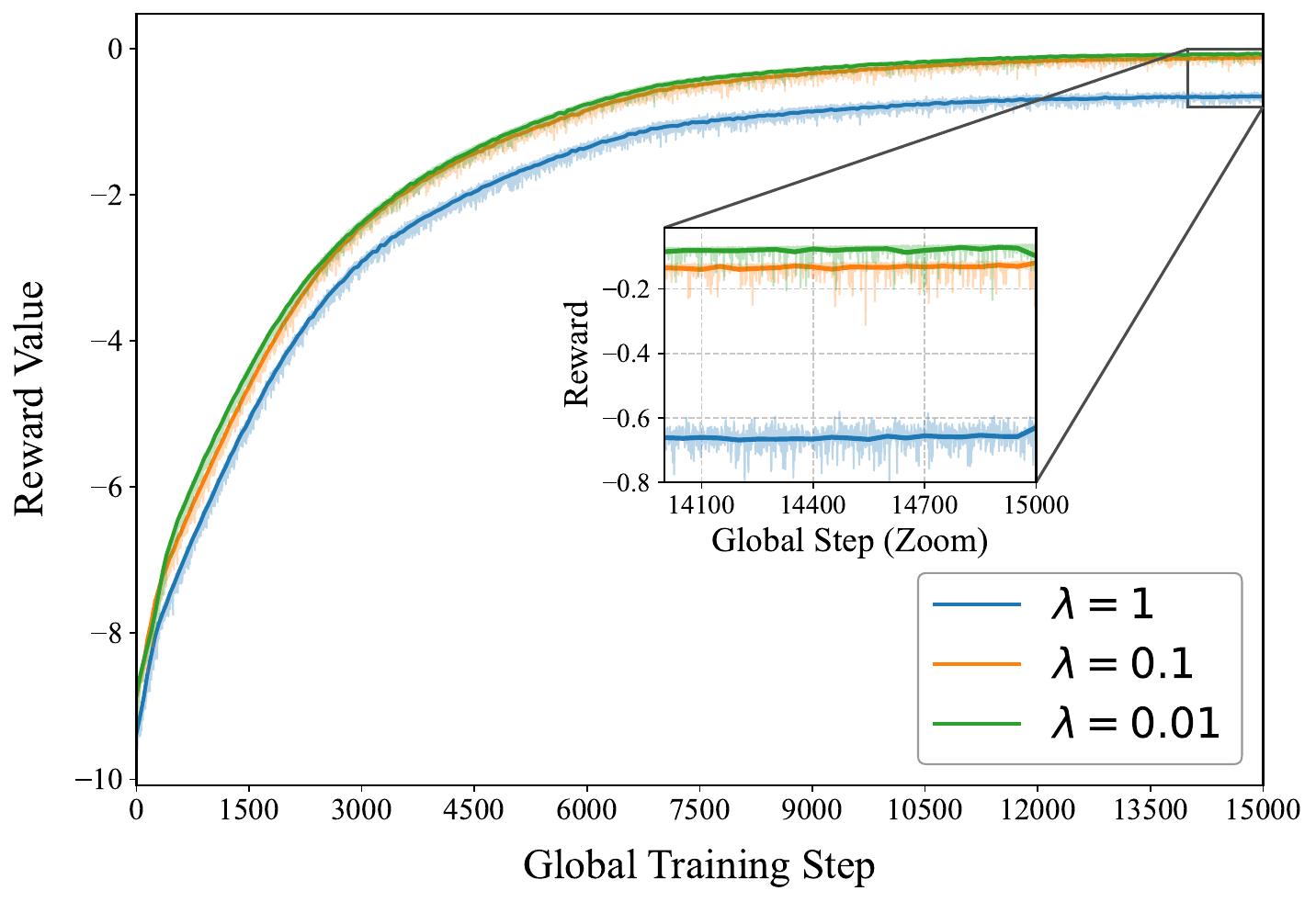}
    \caption{Impact of reward coefficient $\lambda$.}
    \label{fig:lambda}
\end{figure}
\begin{figure}[t]
    \centering
    \includegraphics[width=\linewidth]{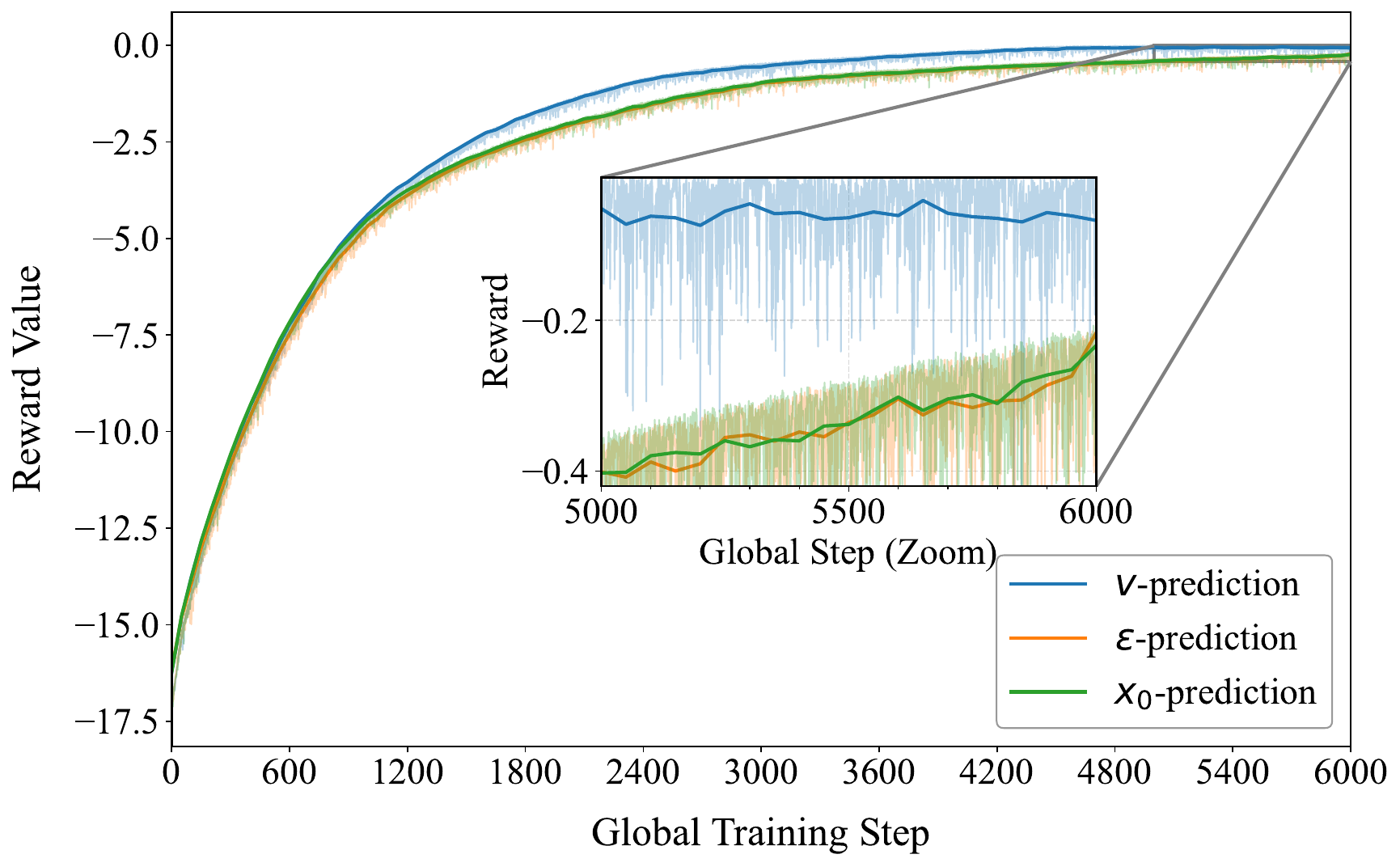}
    \caption{The training reward of policy with different DDIM prediction strategies.}
    \label{fig:ddim}
\end{figure}

\begin{figure}[!ht]
    \centering
\includegraphics[width=\linewidth]{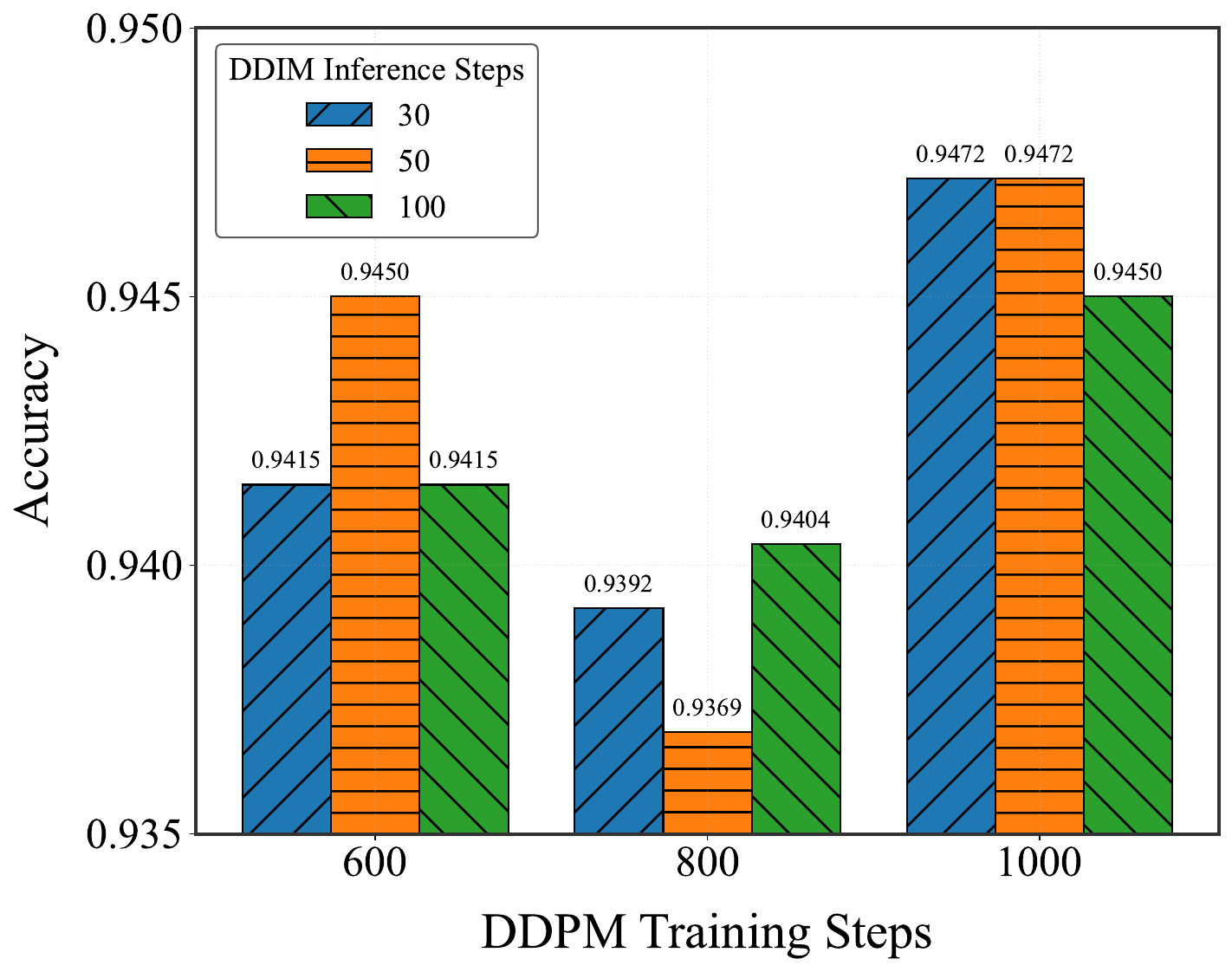}
    \caption{Accuracy performance evaluated under varying DDPM training steps and DDIM Inference steps.  }
    \label{fig:train_infer}
\end{figure}
\begin{figure*}[!htb]
    \centering
    \includegraphics[width=0.85\linewidth]{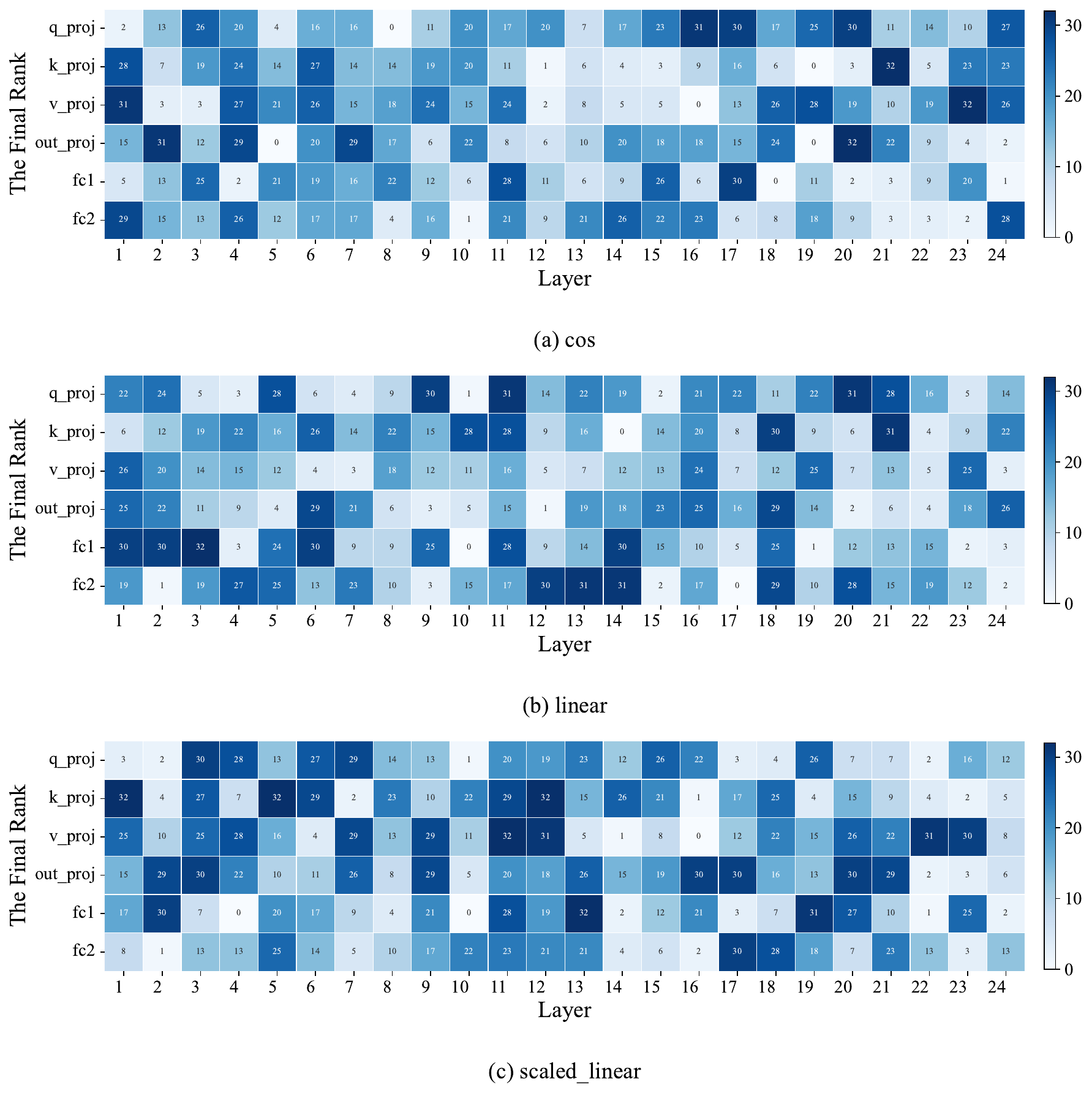}
    \caption{Rank distribution under different $\beta$ schedules.}
    \label{fig:beta}
\end{figure*}
% \begin{table}[tbp]
%     \centering
%     \caption{Prediction Accuracy for Different $\lambda$ Values.}
%     \label{tab:lambda_accuracy_transposed}
%     \begin{tabular}{cccc}
%         \toprule
%         & $\lambda=1$ & $\lambda=0.1$ & $\lambda=0.01$ \\
%         \midrule
%         Accuracy & 95.07\% & 95.41\% & 94.95\% \\
%         \bottomrule
%     \end{tabular}
% \end{table}
% \begin{table}[h]
% \centering
% \caption{Prediction Accuracy for Different Prediction Types.}
% \label{tab:prediction_accuracy}
% \begin{tabular}{lccc}
% \toprule
% & \textbf{$v$-prediction} & \textbf{$\epsilon$-prediction} & \textbf{$x_0$-prediction} \\
% \midrule
% Accuracy & 95.30\% & 95.30\% & 95.07\% \\
% \bottomrule
% \end{tabular}
% \end{table}
\begin{table}[tbp]
    \centering
    \caption{Classification accuracy under different settings.}
    \label{tab:combined_accuracy}
    \begin{tabular}{lccc}
        \toprule
        \multicolumn{4}{c}{\textbf{(a) Accuracy for Different $\lambda$ Values}} \\
        \midrule
        & $\lambda=1$ & $\lambda=0.1$ & $\lambda=0.01$ \\
        Accuracy & 95.07\% & 95.41\% & 94.95\% \\
        \midrule
        \multicolumn{4}{c}{\textbf{(b) Accuracy for Different Prediction Types}} \\
        \midrule
        & \textbf{$\boldsymbol{v}$-prediction} & \textbf{$\boldsymbol{\epsilon}$-prediction} & \textbf{$\boldsymbol{x_0}$-prediction} \\
        Accuracy & 95.30\% & 95.30\% & 95.07\% \\
        \bottomrule
    \end{tabular}
\end{table}

% \begin{figure*}[tb]
%     \centering
%     \includegraphics[width=0.85\linewidth]{Figures/lr_combination.pdf}
%     \caption{Learning rate sensitivity.}
%     \label{fig:lr}
% \end{figure*}

Next, we examine the impact of the reward balancing coefficient $\lambda$, which as defined in Eq. \eqref{eq:reward}, regulates the trade-off between task accuracy and communication cost. Fig. \ref{fig:lambda} shows $\lambda=0.1$ or $\lambda=0.01$ achieves superior, competitive balance, namely higher reward values with faster convergence. Table \ref{tab:combined_accuracy}(a) further confirms the superiority of $\lambda=0.1$ in classification accuracy. On the contrary, a smaller $\lambda$ compromises the balance. % This validates that $\lambda=0.1$ effectively weights task loss and communication cost.

We further consider different DDIM prediction strategies \cite{chi2023diffusion}, each representing a distinct modeling choice. The $\boldsymbol{\epsilon}$-prediction estimates residual noise, directly aligning with the probabilistic structure of the diffusion process for stable and accurate reconstruction. The $\boldsymbol{v}$-prediction models a velocity-like intermediate variable defined as
$
\boldsymbol{v} = \sqrt{\bar{\alpha}_t} \, \boldsymbol{\epsilon} - \sqrt{1 - \bar{\alpha}_t} \, \boldsymbol{x}_0,
$
which combines both noise and clean signal components, offering a balanced view of the denoising dynamics.
 The $\boldsymbol{x_0}$-prediction directly targets the clean data, but this approach is more sensitive to boundary effects in high-dimensional spaces, leading to larger convergence fluctuations. 
Table \ref{tab:combined_accuracy}(b) shows that both $\boldsymbol{\epsilon}$-prediction and $\boldsymbol{v}$-prediction achieve high accuracy, outperforming $\boldsymbol{x_0}$-prediction at a lower value. Similarly, Fig \ref{fig:ddim} shows $\boldsymbol{v}$-prediction obtain higher reward, however, the other two prediction methods also gain a high reward. These results further confirm the advantage of residual noise or intermediate variable modeling over direct data recovery for dynamic rank allocation in diffusion-based policies.

% Learning rate combinations must balance PPO's trust region constraints Eq.~\eqref{eq:ppo_loss} and diffusion's noise estimation. Fig. \ref{fig:lr} shows $\text{lr}_{\text{PPO}}$=$0.0001$ and $\text{lr}_{\text{Diffusion}}$=$0.0005$ yield $95.53$\% accuracy. Larger PPO rates ($\geq0.001$) cause oscillations by violating the clip constraint $[1-\epsilon,1+\epsilon]$, while diffusion tolerates higher rates due to DDIM's deterministic sampling reducing exploration noise. This matches theoretical expectations: PPO requires conservative updates to maintain stability, while diffusion benefits from faster learning of noise patterns.
\begin{table}[t]
\centering
\caption{Total inference time (s) for different inference steps.}
\label{tab:time-600}
\footnotesize
\begin{tabular}{ccc}
\toprule
\textbf{DDPM Training Steps} & \textbf{DDIM Inference Steps} & \textbf{Total Time (s)} \\
\midrule
$600$ & $30$  & $907$  \\
$600$ & $50$  & $1,037$ \\
$600$ & $100$ & $1,101$ \\
\bottomrule
\end{tabular}
\end{table}

Fig. \ref{fig:train_infer} presents the impact of different training and inference steps. It can be observed that the configuration with $1,000$ steps for training and $50$ steps for inference achieves $95.30\%$ accuracy. As shown in Table~\ref{tab:time-600}, under the same number of DDPM training steps, we set different DDIM inference steps to complete the full training of $1,000$ episodes. The comparison indicates that the training time increases significantly as the number of inference steps grows, demonstrating a direct impact of inference steps on the overall training duration. \textcolor{black}{For DDIM inference steps ($T_\text{diff}$), increasing the inference steps from $30$ to $100$ raises the total time by over $20\%$ while bringing only marginal accuracy improvement.} Sufficient training enables the diffusion model to learn noise patterns, while excessive inference steps ($T_\text{diff}\geq 50$) increase latency by $40\%$ with negligible gains.

\textcolor{black}{In addition, we provide comprehensive sensitivity analysis on noise schedule parameters ($\beta$), which govern the diffusion trajectory and final rank allocation behavior.} The $\beta$ schedule in Eq.~\eqref{eq:mu_theta} shapes the rank distribution by controlling noise degradation. Fig. \ref{fig:beta} shows that the scaled-linear schedule outperforms alternatives and assigns higher average ranks to lower layers. For the evaluated OPT-1.3B model, this allocation pattern indicates a greater contribution from lower-layer modules under the selected objective. In addition, the schedule allocates larger ranks to critical modules such as $\mathbf{W}^O$ and $\mathbf{W}^\text{fc1}$, aligning with the findings of AdaLoRA~\cite{zhang2023adalora} on the importance of these components for preserving task performance. This matches the cumulative SNR expression $\bar{\alpha}_t = \prod(1 - \beta_s)$, where slower noise degradation in lower layers preserves essential rank information, while faster degradation in higher layers effectively prunes redundant parameters, thereby improving communication efficiency for OPT-1.3B.

\section{Conclusion \label{Conclusion}}
This paper has presented AirLLM, a novel framework for communication-efficient remote fine-tuning of LLMs. Compared to the static rank allocation in AdaLoRA, AirLLM has integrated PPO with DDIM to dynamically optimize allocated ranks, further affecting the fine-tuned SVD matrices for over-the-air transmission. The PPO and DDIM components are alternately trained in a decoupled manner, where DDIM adopts the CFG strategy to improve controllability and reward alignment. By embedding channel conditions (SNR and bandwidth) and dataset characteristics (lexical entropy and out-of-vocabulary rate) into an RL state space, AirLLM has adaptively balanced model performance and parameter transmission efficiency. Particularly, AirLLM has achieved up to a $0.69$\% accuracy improvement over AdaLoRA while reducing transmitted parameters by $12.5$\% at $r_{\text{max}}=64$. The convergence speed has been accelerated by $20$--$30$\% via DDIM sampling, showcasing the framework's effectiveness in high-dimensional action spaces. These findings have validated AirLLM's superior efficiency in remote fine-tuning of LLMs.
Future work may explore multi-agent collaboration and adaptive noise scheduling to enhance scalability and robustness in extreme channel conditions.

\bibliographystyle{IEEEtran}
\bibliography{reference}
\end{document}